%% file: main.tex
\documentclass[10pt]{article} % For LaTeX2e
\usepackage[accepted]{tmlr}
\input{math_commands.tex}

\usepackage{xcolor}
\definecolor{brickred}{rgb}{0.8, 0.25, 0.33}
\definecolor{color2}{HTML}{009B55}
\definecolor{color3}{HTML}{00A99A}
\usepackage{hyperref}
\hypersetup{colorlinks=true,urlcolor=color3,linkcolor=color3,citecolor=color3}

\usepackage{booktabs}
\usepackage{amsfonts}
\usepackage{nicefrac}
\usepackage{microtype}
\usepackage{wrapfig}
\usepackage{microtype}
\usepackage{graphicx}
\usepackage{placeins}
\usepackage{subcaption}
\usepackage{multirow}
\usepackage{enumitem}
\usepackage{mathtools}
\usepackage{amsthm}
\usepackage{nccmath}
\usepackage{algorithm}
\usepackage[algo2e, ruled]{algorithm2e}
\SetKwComment{Comment}{$\triangleright$\ }{}
\SetKwProg{Init}{Initialize}{}{}
\usepackage[capitalize,noabbrev]{cleveref}

\newcommand{\methodshort}[1]{$\mathtt{ComPEFT}$}
\newcommand{\methodshortbold}[1]{\textbf{ComPEFT}}
\newcommand{\methodlora}[1]{$\mathtt{ComLoRA}$}
\newcommand{\methodia}[1]{$\mathtt{Com(IA)^3}$}
\newcommand{\method}[1]{Compression for Communicating Parameter Efficient Updates via Sparsification and Quantization}

\title{ComPEFT: \method{}}

\author{
Prateek Yadav$^1$\quad 
Leshem Choshen$^{2,3}$\quad
Colin Raffel~$^{4,5}$\quad
Mohit Bansal$^1$\\
$^1$ UNC-Chapel Hill, $^2$ MIT, $^3$ MIT-IBM Watson AI Lab,\\
$^4$ University of Toronto, $^5$ Vector Institue \\
Correspondence Email: \{praty@cs.unc.edu\}
}

\def\month{04}  % Insert correct month for camera-ready version
\def\year{2025} % Insert correct year for camera-ready version
\def\openreview{\url{https://openreview.net/forum?id=CovLQwu611}} % Insert correct link to OpenReview for camera-ready version

\begin{document}

\maketitle

\begin{abstract}
Parameter-efficient fine-tuning (PEFT) enables creation of specialized language models for diverse tasks, resulting in numerous expert modules. In many practical use cases, these expert PEFT modules are integrated into a single model that answers arbitrary queries by routing queries to different experts. However, only a few experts can be kept in GPU memory due to memory constraints. Consequently, expert modules are frequently loaded and offloaded between CPU/GPU memory or disk storage. This frequent swapping dramatically increases communication overhead, leading unacceptable latency and degrading user experience. The large size of modern PEFT modules further exacerbates this latency. For example, QLoRA experts for $\mathtt{65B~LLaMA}$ are $\mathtt{3.2GB}$, making swapping a major communication bottleneck, particularly in memory-constrained environments. To address these issues, we present \methodshort{} (compressed PEFT), a novel method for compressing fine-tuning residuals (task vectors) of PEFT models. Reducing expert PEFT module size effectively addresses both memory and communication limitations, facilitating faster swapping and enabling a higher density of experts within a given memory footprint. \methodshort{} employs sparsification and ternary quantization to reduce PEFT module size without any additional training while preserving or enhancing model performance. Extensive evaluation across $\mathtt{T5}$, $\mathtt{T0}$, and $\mathtt{LLaMA}$-based models with $\mathtt{200M-65B}$ parameters, \methodshort{} achieves compression ratios of $\mathtt{8x-50x}$. Specifically, we show that \methodshort{} improves with scale -- stronger models exhibit higher compressibility and better performance. We show \methodshort{} applied to $\mathtt{LLaMA-65B}$ outperforms QLoRA by $\mathtt{4.16}\%$ on MMLU with a $\mathtt{26x}$ storage size reduction. Additionally, compressed experts produced by \methodshort{} maintain few-shot compositional generalization capabilities, facilitate efficient communication and computation, and exhibit enhanced performance when merged. Lastly, we provide an analysis of different method components, compare \methodshort{} with other PEFT methods, and test its efficacy for compressing full finetuning residual.\footnote{Code is available at \url{https://github.com/prateeky2806/ComPEFT}.}
\end{abstract}

\input{sections/introduction}

\input{sections/method}

\input{sections/experiments}

\input{sections/related_work}

\section{Conclusion}

Our PEFT compression method, \methodshort{}, offers an effective solution to the latency challenges associated with retrieving expert models. By compressing fine-tuning residuals through sparsification and quantization, \methodshort{} achieves high compression ratios and often enhances model performance across various NLP tasks and model sizes. Moreover, it preserves few-shot compositional generalization capabilities, facilitates efficient communication and computation, and demonstrates improved performance when merged with original models. This research contributes valuable insights into the realm of parameter-efficient fine-tuning, addressing both performance and latency concerns.

\bibliography{main}
\bibliographystyle{tmlr}

\appendix
\input{sections/appendix}

\end{document}

%% file: math_commands.tex
\usepackage{amsmath,amsfonts,bm}

\def\eqref#1{equation~\ref{#1}}
\def\1{\bm{1}}

\DeclareMathAlphabet{\mathsfit}{\encodingdefault}{\sfdefault}{m}{sl}
\SetMathAlphabet{\mathsfit}{bold}{\encodingdefault}{\sfdefault}{bx}{n}

%% file: sections/introduction.tex
\section{Introduction}
\label{sec:introduction}

Parameter-efficient fine-tuning (PEFT)~\citep{houlsby2019parameter_adapter,karimi2021compacter} methods like LoRA~\citep{hu2021lora} and (IA)$^3$~\citep{liu2022tfew} efficiently adapt language models by learning only a few new parameters. QLoRA~\citep{dettmers2023qlora} further reduces memory needs by using 4-bit quantization for the base model. This combined efficiency has fueled a surge in specialized models for diverse tasks such as multimodal understanding~\citep{zhang2023llama_adapter}, multilingual applications~\citep{yang2023bigtrans}, and expert systems for math~\citep{luo2023wizardmath} or coding~\citep{luo2023wizardcoder}. Platforms like HuggingFace Hub~\citep{wolf2019huggingface} now host a rapidly growing collection of these expert PEFT models.

\begin{figure}[t!]
\centering
    \includegraphics[width=0.95\linewidth]{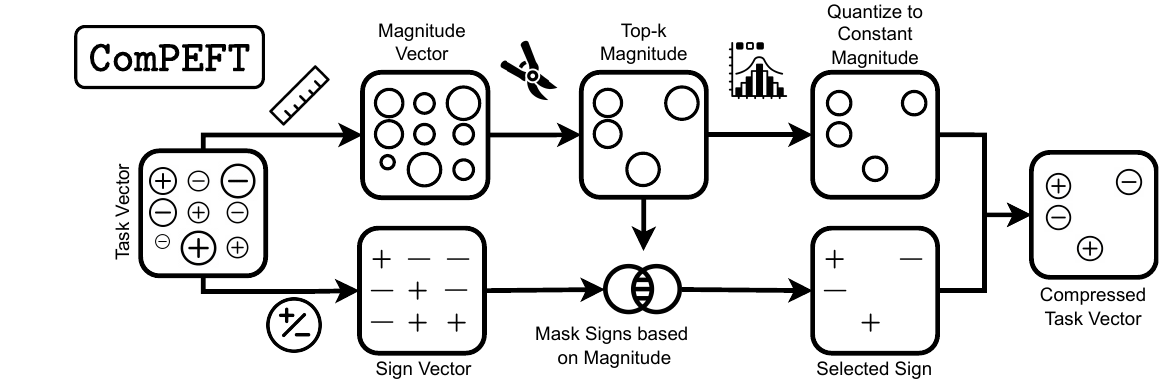}
    % \vspace{-15pt}
    \captionsetup{type=figure}
    \caption{\label{fig:compeft}\methodshort{} without any additional training compresses PEFT modules while preserving or enhancing model performance.
    }
    \vspace{-10pt}
\end{figure}

Serving these expert PEFT models has different strategies. LoRA, for instance, allows merging expert PEFT modules into the base model for single-expert low-latency inference and proposes expert swapping for sequential multi-expert serving. While efficient for single and sequential multi-expert serving, these methods become slow for concurrent multi-expert serving because swapping reduces throughput and increases latency~\citep{sheng2023slora}. Furthermore, LoRA's approaches don't fully utilize available GPU memory for a larger number of experts. For efficient high-throughput concurrent serving, separating base model and adapter computations is crucial, as multiple experts can then share the base.
\begin{wrapfigure}{r}{0.5\textwidth}
\centering
    \includegraphics[width=\linewidth]{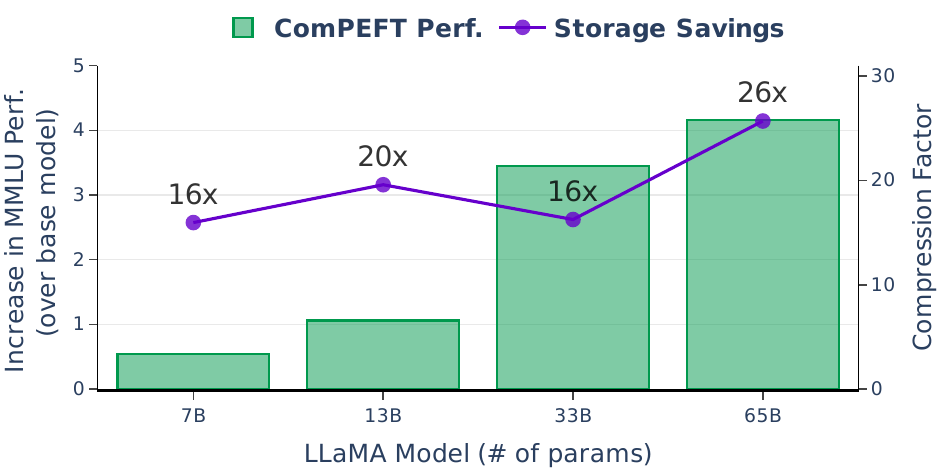}
    \vspace{-15pt}
    \captionsetup{type=figure}
    \caption{\label{fig:intro} \textbf{\methodshort{} improves performance with larger base models while compressing models significantly.} \textbf{Left Axis:} Improvement of MMLU performance over the corresponding base model. \textbf{Right Axis:} Compression factor achieved by \methodshort{} compared to the original checkpoint.
    }
\end{wrapfigure}
This enables efficient base model batching, but directly batching expert PEFT modules remains challenging. Serving numerous experts demands efficient memory management. Limited GPU memory necessitates storing experts off-GPU and dynamically fetching them when needed. This dynamic loading of large expert modules leads to communication overhead and unacceptable latency, degrading user experience~\citep{sheng2023slora}. Communication bottlenecks extend beyond concurrent multi-expert serving. Techniques like Model merging~\citep{ilharco2023editing,yadav2023ties-merging}, Model MoErging~\citep{yadav2024survey}, and compositional generalization (CG)~\citep{huang2023lorahub,muqeeth2024phatgoose} also require dynamically retrieving expert PEFT modules from cloud/disk/cache into GPU memory based on input queries to dynamically merge or route through experts for improved generalization. Consequently, these methods also face communication challenges. For example, a $\mathtt{3.2~GB}$ QLoRA adapter for $\mathtt{LLaMA-65B}$ (comparable to a full T5-Large model~\citep{raffel2019exploring}) can make frequent swapping a bottleneck, especially under memory constraints. Therefore reducing expert PEFT module size solves both memory and communication issues by facilitating both faster swapping and increased expert density within a given memory footprint.

To address these issues, we introduce our \methodshort{} (compressed PEFT) method that compresses fine-tuning residuals -- i.e., task vectors -- by exploiting their inherent redundancies~\citep{yadav2023ties-merging}. The task vectors represent the learned changes to the model's parameters during fine-tuning for a specific task. \methodshort{} achieves this compression through a two-step process. First, it applies sparsification, aggressively sets a large portion of the values within the PEFT task vector to zero. This step is based on the observation that many values in task vectors are close to zero and contribute minimally to the expert's behavior~\citep{yadav2023ties-merging}. Second, for the remaining non-zero values, \methodshort{} employs ternary quantization. Instead of storing these values at full precision, it represents their magnitudes using a single, shared scalar constant, and their signs ($+1,~-1,~\text{or}~0$). This results in task vectors with sparse ternary weights, drastically reducing their size (see Figure~\ref{fig:compeft}). \methodshort{} shares similarities with the Sparse Ternary Compression (STC,~\citealp{sattler2019robust_stc}) method used in federated learning, however, there are notable differences. Unlike STC, \methodshort{} retains high-performance without the need for additional training. This is in stark contrast to directly applying STC to task vectors, which we find leads to performance degradation. Remarkably, \methodshort{} can often restore and even surpass the original fine-tuned performance by carefully calibrating the magnitude of the shared scalar constant used in ternary quantization. Additionally, we demonstrate this effectiveness not only for PEFT modules but also for fully fine-tuned models. Moreover, we observe a beneficial trend that the optimal magnitude of this scalar constant becomes consistent across different tasks for larger models ($\mathtt{\geq 13B}$). This eliminates the need for task-specific tuning and simplifies the practical deployment of \methodshort{} at scale, further facilitating reduced latency during model serving. Finally, the \methodshort{} compression enables more efficient operations on task vectors that can facilitate faster merging of models and compute their similarity.

We perform comprehensive experiments for \methodshort{} to evaluate: (1) the performance of the compressed model on its original tasks, (2) the number of bits needed to store the models, (3) the mergeability and composability of the compressed checkpoints, and (4) how \methodshort{} compares to other existing PEFT methods. We performed experiments with $\mathtt{T5}$~\citep{colin2020exploring}, $\mathtt{T0}$~\citep{sanh2021multitask}, $\mathtt{LLaMA}$~\citep{touvron2023llama}, and $\mathtt{LLaMA2}$~\citep{touvron2023llama2} as the base models with model sizes ranging from $\mathtt{200M-70B}$ parameters. We found that in most cases \methodshort{} can provide compression of $\mathtt{8x-50x}$ (compared to $\mathtt{16-bit}$ precision checkpoints) while performing similarly or better than the uncompressed models. Additionally, we note a surprising finding that as the base model gets bigger, their task vectors become more compressible and these compressed checkpoints significantly outperform the original uncompressed checkpoints. Specifically, as shown in Figure~\ref{fig:intro},  \methodshort{} leads to an improvement of $\mathtt{0.54}\%$, $\mathtt{1.06}\%$, $\mathtt{3.44}\%$, and $\mathtt{4.16}\%$ on MMLU for QLoRA trained on $\mathtt{LLaMA}$ $\mathtt{7B}$, $\mathtt{13B}$, $\mathtt{33B}$, and $\mathtt{65B}$ parameter models, respectively, while compressing model by $\mathtt{16x}-\mathtt{26x}$. Beyond performance and size, we demonstrate that \methodshort{} provides order-of-magnitude reductions in model transmission and loading latency, directly addressing communication bottlenecks in expert model serving. 
% Moreover, the compressed models from \methodshort{} lead to better-merged models and outperform strong baselines like Task Arithmetic~\citep{ilharco2023editing} and TIES-Merging~\citep{yadav2023ties-merging} in $\mathtt{9/12}$ settings and lead to an improvement of $\mathtt{1.4}\%$ on average across all these settings. 
In addition, we show that (1) the compressed models from \methodshort{} lead to better-merged models; (2) for few-shot compositional generalization (CG)~\citep{huang2023lorahub}, \methodshort{} checkpoints lead to similar performance on BBH~\citep{suzgun2022challenging_bbh}; (3) \methodshort{} applied to LoRA and (IA)$^3$ is Pareto-optimal in terms of storage costs vs. performance compared to a wide range of existing; (4) the importance of the components of \methodshort{} in an ablation study, and (5) the effect of sparsity and scaling on performance. We further show that \methodshort{}'s benefits extend beyond PEFT, effectively compressing fully fine-tuned models with minimal degradation. Our results and analysis establish \methodshort{} as an effective method for compressing task vectors. In summary, our contributions are:
\begin{enumerate}
    \item \methodshort{} demonstrates that even efficient PEFT modules can be drastically compressed ($\mathtt{8x-50x}$) via sparsification and quantization without performance loss, suggesting PEFT modules contain significant redundancy.
    
    \item The reduced size of \methodshort{} checkpoints ($\mathtt{8x-50x}$ smaller) enables serving larger models or significantly more expert PEFT modules given fixed resources.
    
    \item \methodshort{}'s smaller size reduces communication overhead during dynamic retrieval and loading/offloading, leading to improved latency in practical serving systems.
    
\end{enumerate}

%% file: sections/method.tex
\section{\texorpdfstring{\methodshort{}}{methodshort}: \method{}}
\label{sec:method}

As discussed in the introduction (\S\ref{sec:introduction}), serving numerous expert PEFT modules suffer from communication and memory bottlenecks which can be alleviated by compressing the experts. This section details \methodshort{}, our method to compress experts by targeting their fine-tuning residuals. Given a pre-trained model like $\mathtt{LLaMA}$~\citep{touvron2023llama} or $\mathtt{T5}$~\citep{raffel2019exploring}, we can create an expert model for specific task $t$ by either finetuning all model parameters or using a parameter-efficient fine-tuning (PEFT) approach such as (IA)$^3$~\citep{liu2022tfew} or LoRA~\citep{hu2021lora}. In both scenarios, we represent the trainable parameters as $\theta$, initialized as $\theta_\mathtt{init}$, which, upon fine-tuning, become $\theta_\mathtt{ft}$. This work assumes access to the initial model parameters $\theta_{\mathtt{init}}$ and the fine-tuned model parameters, $\theta_\mathtt{ft}$. Our work focuses on two key objectives: (1) to achieve extreme compression of parameter updates for efficient and low-latency communication of expert models, and (2) to gain insights into the inherent compressibility of these updates, suggesting a lower intrinsic dimensionality of learned task-specific knowledge.

For a given task $t$, we first represent the parameter updates from fine-tuning as a task vector $\tau_\mathtt{t} = \theta_{\mathtt{ft}} - \theta_{\mathtt{init}}$. This task vector encapsulates the changes applied to the base model parameters to specialize it for task $t$. To effectively compress these task vectors, we decompose each $\tau_\mathtt{t}$ into direction and magnitude components. This decomposition allows us to treat the direction of parameter updates and their magnitude separately, enabling us to apply distinct compression strategies optimized for each component. We decompose the task vector $\tau_\mathtt{t}$ into a direction (sign) vector 
$\gamma_\mathtt{t} \in \mathbb{R}^d$ and a magnitude vector $\mu_\mathtt{t} \in \mathbb{R}^d$. Formally, the direction vector $\gamma_\mathtt{t} = \text{sgn}(\tau_\mathtt{t})$ captures the sign of each parameter (+$\mathtt{1}$, $\mathtt{0}$ or -$\mathtt{1}$), while the magnitude vector $\mu_\mathtt{t}$ is $\mu_\mathtt{t} = |\tau_\mathtt{t}|$ captures the absolute magnitude. This decomposition allows us to express the task vector as the Hadamard product: $\tau_\mathtt{t} = \gamma_\mathtt{t} \odot \mu_\mathtt{t}$. Based on the intuition from~\citet{yadav2023ties-merging}, that the direction of parameter updates is crucial for task adaptation, while lower magnitude updates are redundant. \methodshort{} achieves high compression by sparsifying the direction vector $\gamma_\mathtt{t}$ and quantizing the magnitude vector $\mu_\mathtt{t}$ to a single scalar.
% The effectiveness of this approach is empirically validated across diverse models and tasks in our main results (\S\ref{sec:main_results} and \S\ref{sec:additional_results}), where \methodshort{} consistently achieves significant compression with minimal performance degradation, and even Pareto-optimal performance compared to other PEFT methods (\S\ref{sec:pareto_peft}).

\setlength{\intextsep}{2pt}
\setlength{\columnsep}{20pt}
\begin{wrapfigure}{r}{0.5\textwidth}
\begin{minipage}{\linewidth}
\vspace{-23pt}
\begin{algorithm}[H]
\caption{\label{alg:main} \methodshort{} Compression Procedure.}
\DontPrintSemicolon
\KwIn{Task vector $\tau_{\mathtt{t}}$, $k$, and a scaling value $\alpha$.}
\KwOut{Compressed task vector $\tilde{\tau}_\mathtt{t}$}

% \Comment{\scriptsize Step 1: Decompose.}

$\gamma_\mathtt{t} \leftarrow sgn(\tau_\mathtt{t})$

$\mu_\mathtt{t} \leftarrow |\tau_\mathtt{t}|$

\Comment{\scriptsize Step 1: Sparsify.}
$\tilde{\tau}_\mathtt{t} \leftarrow \text{keep\_topk\_reset\_rest\_to\_zero}(\gamma_\mathtt{t}, \mu_\mathtt{t}, k)$

\Comment{\scriptsize Step 2: Quantize Magnitudes to scalar.}

$\tilde{\tau}_\mathtt{t} = \alpha * \sigma(\tau_\mathtt{t}) * \tilde{\gamma}_\mathtt{t}$ 

\Return{$\tilde{\tau}_\mathtt{t}$} 
\end{algorithm}
\end{minipage}
\end{wrapfigure}

\subsection{Steps in \texorpdfstring{\methodshort{}}{methodshort} }

To reconstruct an expert model for task $t$, we only need to communicate the compressed update over the base pre-trained model, which is represented by the task vector $\tau_\mathtt{t}$. As described earlier, we decompose this task vector into a direction vector $\gamma_\mathtt{t}$ and a magnitude vector $\mu_\mathtt{t}$. Given this decomposition, \methodshort{} compresses the task vector through two key steps of sparsification and quantization. Refer to Algorithm~\ref{alg:main} and Figure~\ref{fig:compeft}.

\begin{enumerate}[leftmargin=2em,labelsep=0.5em]

    \item \textbf{Sparsify:} We sparsify the direction vector $\gamma_\mathtt{t}$ by retaining only the signs of the parameters corresponding to the top-$k\%$ largest magnitudes in $\mu_\mathtt{t}$, and setting the signs of the remaining $(1-k)\%$ parameters to zero. Following ~\citet{yadav2023ties-merging}, we select the top-$k\%$ parameters based on their magnitude in $\mu_\mathtt{t}$ as larger magnitude updates generally represent more significant parameter changes learned during fine-tuning. By preserving the signs of these largest magnitude updates and zeroing out the rest, we aim to retain the most critical directional information for each task. Formally, the sparsified direction vector, $\tilde{\gamma}_\mathtt{t} = \gamma_\mathtt{t} \odot \text{top-}k(\mu_\mathtt{t})$, where top-$k$(.) is applied elementwise and returns $\mathtt{1}$ for indices with the $\mathtt{top-k\%}$ magnitude values and $\mathtt{0}$ otherwise. The parameter $k$ is referred to as the "\textit{density}", and $1-k$ as the \textit{sparsity}.

    \item \textbf{Quantize Magnitudes:} We then quantize the magnitude vector $\mu_\mathtt{t}$ to a single scalar value. Specifically, we define the compressed task vector $\tilde{\tau}_\mathtt{t} \in \mathbb{R}^d$ as $\tilde{\tau}_\mathtt{t} = \alpha * \sigma(\tau_\mathtt{t}) * \tilde{\gamma}_\mathtt{t}$. Here, $\sigma(\tau_\mathtt{t}) \in \mathbb{R}$ is the standard deviation of the original task vector $\tau_\mathtt{t}$, and $\alpha \in \mathbb{R}$ is a scaling hyper-parameter. We utilize the standard deviation of the original task vector as a scaling factor to normalize the magnitude, which helps to make the optimal $\alpha$ value more consistent across different tasks and models. Refer to Appendix~\ref{app:std} for more discussion. The scaling factor $\alpha$ is then chosen by evaluating performance on a small validation set; importantly, $\alpha$ is the only parameter tuned during this process. We observe that this simple scalar scaling is sufficient to effectively mitigate any performance loss from sparsification and ternary quantization. This contrasts with many model pruning methods that require computationally expensive retraining after sparsification to recover performance.
    % 
    % The importance of both the sparsification and the scalar quantization steps, along with the tuned scaling factor $\alpha$, is further validated by our ablation studies (Figure \ref{fig:ablations}), which demonstrate that \methodshort{} consistently outperforms variants with ablated components.
    
\end{enumerate}

\subsection{Efficient Storage of \texorpdfstring{\methodshort{}}{methodshort} Models}
\label{sec:method_storage}

\methodshort{}'s compression strategy directly addresses the communication bottleneck and latency concerns highlighted in \S\ref{sec:introduction} by significantly reducing the storage footprint of expert PEFT modules. This section details the storage efficiency gains and discusses practical encoding schemes for efficient communication and computation.

\textbf{Entropy of the Sparsified Task Vector.}  A typical task vector $\tau_\mathtt{t}$ in $\mathrm{bfloat16}$ or $\mathrm{fp16}$ format requires $\mathtt{16*d~bits}$ for memory/storage. Assuming uniform value distribution, its entropy is also $\mathbb{H}_{\mathtt{dense}} = 16*d~\mathtt{bits}$. \methodshort{}, however, represents the compressed task vector $\tilde{\tau}_\mathtt{t}$ using a sparse ternary sign vector (values $\in \{-1, 0, +1\}$) and a single 16-bit scalar value ($\alpha * \sigma(\tau_\mathtt{t}) \in \mathbb{R}$). Assuming the signs of the nonzero entries of $\tilde{\tau}_\mathtt{t}$ are uniformly distributed, the ternarization step reduces the entropy of the update to $\mathbb{H}_{\mathtt{ComPEFT}} = - ((1-k) \log_2(1-k) + k \log_2(\frac{k}{2})) * d + 16 ~\mathtt{bits}$, where $k$ is the density of the update. At a density level of $k=0.05$, the resultant update has $95\%$ of the values as $0$ and the entropy is $0.34*d+16~\mathtt{bits}$. Hence, with a perfect encoding-decoding scheme and 95\% sparsity, our \methodshort{} can reduce the number of bits per parameter from $16~\mathtt{bits}$ to approximately $0.34~\mathtt{bits}$ which is a $47\mathtt{x}$ improvement in communication and storage costs. We now discuss two practical encoding schemes to realize these savings.

\textbf{Optimal Compression via Golomb Coding.} For maximal compression in communication and storage, Golomb coding is effective. This near-entropy method suits geometrically distributed data, like distances in sparse task vectors~\citep{strom2015scalable,sattler2019robust_stc}. Using Golomb coding~\citep{golomb1966run}, we communicate the locations of non-zero elements with an additional bit indicating each element's sign, achieving near-optimal compression. This approach needs a total of $ - ((1-k) \log_2(1-k) + k \log_2(\frac{k}{2})) * d + 16 ~\mathtt{bits}$ for storage, and its average bits per position $\bar{\texttt{b}}_{pos}$, the calculation of which is detailed in the footnote below\footnote{Similar to~\citet{strom2015scalable,sattler2019robust_stc,sattler2019sparse_sbc}, the average bits per position $\bar{\texttt{b}}_{pos}$ is calculated as follows:
$\bar{\texttt{b}}_{pos} = \mathbf{b}^*+\frac{1}{1-(1-p)^{2^{\mathbf{b}^*}}},$
with $\mathbf{b}^*=1+\lfloor \log_2(\frac{\log(\phi-1)}{\log(1-p)})\rfloor$ and $\phi=\frac{\sqrt{5}+1}{2}$ being the golden ratio.}. Unless otherwise specified, storage costs reported in our experiments assume Golomb coding.

\textbf{Efficient Computation and Communication via Two Binary Vectors.} 
Alternatively, for scenarios prioritizing computational efficiency, \methodshort{} compressed task vector $\tilde{\tau}_\mathtt{t}$ can be represented using two binary masks, one signifying positive values and the other signifying negative values. Formally, we need to communicate $\tilde{\tau}_\mathtt{t}^+ = (\tilde{\tau}_\mathtt{t}$ == $+1) \in \mathbb{R}^d$ and $\tilde{\tau}_\mathtt{t}^- = (\tilde{\tau}_\mathtt{t}$ == $-1) \in \mathbb{R}^d$, and the scalar constant $\alpha * \sigma(\tau_\mathtt{t})$. Each binary mask needs $\mathtt{1~bit/parameter}$, resulting in $\mathtt{2*d + 16~bits}$ for communicating the update. Note that this requires strictly more storage than the Golomb-based encoding described above because $ - ((1-k) \log_2(1-k) + k \log_2(\frac{k}{2})) < 2$. However, sparse ternary vectors allow for efficient matrix operations. For example, to efficiently compute the distance between $\tilde{\tau}_\mathtt{t_1}$ and $\tilde{\tau}_\mathtt{t_2}$, we can do an $\mathtt{XOR}$ ($\oplus$) followed by a $\mathtt{POPCNT}$ for each group of 64 parameters (i.e. two machine instructions on a 64-bit architecture) twice, once for the positive and once for the negative masks. The dot product can also be calculated by using bitwise $\mathtt{AND}$ operations to calculate positive contributions (both vectors have $+1$ or $-1$) and negative contributions (one vector has $+1$, the other $-1$). The final dot product is the difference between the sum of these contributions. Similarly, other operations such as addition can also be made faster, which could reduce the time when merging models. Thus, \methodshort{} offers flexibility to use Golomb coding for optimal storage, and binary vectors for efficient computation.

%% file: sections/experiments.tex
\section{Main Results}
\label{sec:main_results}

\subsection{Compressing QLoRA Trained on LLaMA Models}
\label{sec:peft_llama}

\textbf{Experimental Setup.} We first explore the utility of \methodshort{} in the setting of training QLoRA adapters~\citep{dettmers2023qlora} for the $\mathtt{LLaMA}$ models~\citep{touvron2023llama} with $\mathtt{7B}$, $\mathtt{13B}$, $\mathtt{33B}$, and $\mathtt{65B}$ parameters. We follow the experimental setting from the QLoRA paper~\citep{dettmers2023qlora} and experiment with $\mathtt{8}$ recent instruction-following datasets that are diverse in terms of languages and dataset sizes. This collection includes datasets generated by language models (Alpaca~\citep{alpaca}, self-instruct~\citep{wang2022self}, and unnatural-instructions~\citep{honovich2022unnatural}), a multitask dataset (FLAN-v2~\citep{chung2022scaling}), two datasets created via human annotation and feedback (OASST1~\citep{kopf2023openassistant} and HH-RLHF~\citep{bai2022training}), and two hybrid datasets (Chip2~\citep{laion2023} and Longform~\citep{koksal2023longform}). For each of these datasets, we reuse the checkpoints released with the QLoRA paper\footnote{\href{https://huggingface.co/timdettmers?search\_models=qlora}{https://huggingface.co/timdettmers?search\_models=qlora}} to perform compression using Algorithm~\ref{alg:main} and then evaluate the 5-shot performance of the compressed QLoRA module on the MMLU benchmark~\citep{hendrycks2020measuring_mmlu}. To ensure the generalizability of \methodshort{}, we extend our evaluation to $\mathtt{LLaMA2-70B}$ model.
In all experiments, we sweep both $\alpha$ and $k$ in the following ranges, $k \in \{5, 10, 20, 30, 50\}$ and $\alpha \in \{0.5, 1, 2, 3, 4, 5, 6, 8, 10\}$ and report the storage size based on the entropy of \methodshort{} as specified in \S\ref{sec:method_storage}. We find that at any given value of $k$, you can achieve good performance (see \S~\ref{sec:spar_scale}).
We used a single 48GB NVIDIA A6000 GPU for these experiments.

\begin{table}[t!]
\vspace{-10pt}
\centering
\captionsetup{type=table}
\caption{\label{tab:qlora} \textbf{Performance improvement from \methodshort{} increases as models get bigger.} We present the performance ${\text{(storage size in GB)}}$ on the MMLU Test for the original and compressed QLoRA models. For $\mathtt{LLaMA-65B}$, \methodshort{} leads to a $\mathtt{4.16\%}$ improvement while being $\mathtt{26x}$ smaller.}
\vspace{-5pt}
\resizebox{\linewidth}{!}{  
\begin{tabular}{lcccccccc}
\toprule

\textbf{Model Size ($\rightarrow$)} & \multicolumn{2}{c}{\textbf{7B}}  & \multicolumn{2}{c}{\textbf{13B}} & \multicolumn{2}{c}{\textbf{33B}} & \multicolumn{2}{c}{\textbf{65B}} \\
\cmidrule(lr){2-3} \cmidrule(lr){4-5} \cmidrule(lr){6-7} \cmidrule(lr){8-9}

\textbf{Dataset ($\downarrow$)} & \textbf{Original} & \methodshortbold{} & \textbf{Original} & \methodshortbold{} & \textbf{Original} & \methodshortbold{} & \textbf{Original} & \methodshortbold{} \\
\midrule

% $\mathtt{LLaMA~ no~ tuning}$ & 35.1 & - & 46.9 & - & 57.8 & - & 63.4 & - \\
% \midrule

\textbf{Self-Instruct} & 36.45$_{(0.3)}$ & \textbf{37.72}$_{(0.03)}$ & 36.20$_{(0.47)}$ & \textbf{45.15}$_{(0.01)}$ & 50.98$_{(0.91)}$ & \textbf{57.02}$_{(0.02)}$ & 55.34$_{(1.49)}$ & \textbf{63.43}$_{(0.03)}$  \\
\textbf{Longform} & 34.37$_{(0.3)}$ & \textbf{35.48}$_{(0.02)}$ & 45.70$_{(0.47)}$ & \textbf{46.80}$_{(0.02)}$ & 54.60$_{(0.91)}$ & \textbf{57.07}$_{(0.07)}$ & 59.49$_{(1.49)}$ & \textbf{62.95}$_{(0.05)}$  \\
\textbf{Chip2} & 34.88$_{(0.3)}$ & \textbf{36.11}$_{(0.02)}$ & 44.19$_{(0.47)}$ & \textbf{45.06}$_{(0.03)}$ & 51.72$_{(0.91)}$ & \textbf{56.43}$_{(0.03)}$ & 57.30$_{(1.49)}$ & \textbf{63.32}$_{(0.05)}$  \\
\textbf{HH-RLHF} & \textbf{35.52}$_{(0.3)}$ & 35.30$_{(0.01)}$ & 44.66$_{(0.47)}$ & \textbf{44.99}$_{(0.01)}$ & 53.41$_{(0.91)}$ & \textbf{56.97}$_{(0.07)}$ & 58.79$_{(1.49)}$ & \textbf{63.42}$_{(0.05)}$  \\
\textbf{Unnatural Instruct} & \textbf{42.14}$_{(0.3)}$ & 41.82$_{(0.02)}$ & \textbf{48.98}$_{(0.47)}$ & 48.42$_{(0.03)}$ & 56.65$_{(0.91)}$ & \textbf{58.07}$_{(0.09)}$ & 59.50$_{(1.49)}$ & \textbf{63.30}$_{(0.03)}$  \\
\textbf{Guanaco (OASST1)} & 35.02$_{(0.3)}$ & \textbf{36.31}$_{(0.01)}$ & \textbf{48.50}$_{(0.47)}$ & 47.10$_{(0.03)}$ & 55.51$_{(0.91)}$ & \textbf{57.55}$_{(0.05)}$ & 60.67$_{(1.49)}$ & \textbf{63.25}$_{(0.09)}$  \\
\textbf{Alpaca} & \textbf{40.72}$_{(0.3)}$ & 39.95$_{(0.02)}$ & \textbf{49.53}$_{(0.47)}$ & 48.41$_{(0.03)}$ & 53.66$_{(0.91)}$ & \textbf{57.68}$_{(0.05)}$ & 60.51$_{(1.49)}$ & \textbf{63.28}$_{(0.05)}$  \\
\textbf{FLAN v2} & 43.97$_{(0.3)}$ & \textbf{44.70}$_{(0.02)}$ & 50.45$_{(0.47)}$ & \textbf{50.76}$_{(0.03)}$ & 56.67$_{(0.91)}$ & \textbf{60.01}$_{(0.07)}$ & 62.72$_{(1.49)}$ & \textbf{64.61}$_{(0.11)}$  \\

\midrule
\textbf{Average} & 37.88$_{(0.3)}$ & 38.42$_{(0.0188)}$ & 46.03$_{(0.47)}$ & 47.09$_{(0.024)}$ & 54.15$_{(0.91)}$ & 57.60$_{(0.056)}$ & 59.29$_{(1.49)}$ & 63.45$_{(0.058)}$ \\
\textbf{Increase/Comp.} & $-$ & \textcolor{color2}{+0.54} / \textcolor{color2}{$\mathtt{16x}$} & $-$ & \textcolor{color2}{+1.06} / \textcolor{color2}{$\mathtt{20x}$} & $-$ & \textcolor{color2}{+3.44} / \textcolor{color2}{$\mathtt{16x}$} & $-$ & \textcolor{color2}{+4.16} / \textcolor{color2}{$\mathtt{26x}$} \\

\bottomrule
\end{tabular}
}
\end{table}

\begin{figure}[t!]
  \vspace{-5pt}
  \centering
  \begin{minipage}{0.49\linewidth}
    \centering
    \captionsetup{type=table}
    \caption{\label{tab:llama2} \textbf{LLaMA2-70B Results.} Mirroring our main findings, \methodshort{} improves average performance (by $\mathtt{1.69\%}$) on $\mathtt{LLaMA2-70B}$, notably by $\mathtt{4.82\%}$ on Self-Instruct.
}
    
    \begin{tabular}{lcc}
    \toprule
    \textbf{Dataset ($\downarrow$)} & \textbf{Original} & \methodshortbold{} \\ 
    \midrule
    \textbf{Alpaca}        & 67.13    & 67.56 \textcolor{color2}{(+0.43)}   \\
    \textbf{Chip2}         & 65.18    & 67.00 \textcolor{color2}{(+1.82)}      \\
    \textbf{Longform}      & 67.63    & 68.50 \textcolor{color2}{(+0.86)}    \\
    \textbf{Guanaco}       & 66.89    & 67.39 \textcolor{color2}{(+0.5)}   \\
    \textbf{Self-Instruct} & 62.36    & 67.18 \textcolor{color2}{(+4.82)}   \\
    \midrule
    \textbf{Average}       & 65.84   & 67.53 \textcolor{color2}{(+1.69)}  \\ \bottomrule
    \end{tabular}
  \end{minipage}
  \hfill
    \begin{minipage}{0.49\linewidth}
        \centering
        \captionsetup{type=table}
        \caption{\label{tab:main_goodzs_peft_test} \textbf{\methodshort{} can compress smaller model with minimal performance loss.} Test set performance$_\text{(Storage Size in MB)}$ averaged over seven GLUE tasks when compressing (IA)$^3$ and LoRA modules on different base models.}
        \vspace{-0.2cm}
        \resizebox{\linewidth}{!}{  
        \begin{tabular}{lcccc}
        \toprule
        
        \textbf{PEFT ($\downarrow$)} & \textbf{Method ($\downarrow$)} &  \textbf{T5-Base}  & \textbf{T5-Large} & \textbf{T0-3B} \\
        \midrule
        
        \multirow{3}{*}{\textbf{(IA)}$^\textbf{3}$} & \textbf{Original} & 81.3$_{(0.25)}$  & 86.2$_{(0.66)}$ & 89.3$_{(1.03)}$  \\
        & \methodshortbold{} & 80.0$_{(0.01)}$ & 85.9$_{(0.04)}$ & 88.4$_{(0.06)}$  \\
        \cmidrule(lr){2-5}
        & \textbf{Improvement} & \textcolor{brickred}{-1.3} / \textcolor{color2}{$\mathtt{25x}$}  & \textcolor{brickred}{-0.3} / \textcolor{color2}{$\mathtt{16x}$} & \textcolor{brickred}{-0.9} / \textcolor{color2}{$\mathtt{17x}$} \\
        \midrule
        \multirow{3}{*}{\textbf{LoRA}} & \textbf{Original} &	79.2$_{(6.19)}$ & 84.5$_{(16.50)}$ & 89.5$_{(33.75)}$ \\
        & \methodshortbold{} & 78.1$_{(0.35)}$ & 84.6$_{(1.37)}$ & 89.5$_{(2.60)}$ \\
        \cmidrule(lr){2-5}
        & \textbf{Improvement} & \textcolor{brickred}{-1.1} / \textcolor{color2}{$\mathtt{17x}$} & \textcolor{color2}{+0.1} / \textcolor{color2}{$\mathtt{12x}$} &  0.0 / \textcolor{color2}{$\mathtt{13x}$} \\
        \bottomrule
        \end{tabular}
        }
  \end{minipage}
  \vspace{-10pt}
\end{figure}

\textbf{Outcomes.} In Table~\ref{tab:qlora}, we provide results for all the task and model size combinations, comparing the performance of the \methodshort{} checkpoints and the original QLoRA checkpoints along with (in subscripts) the storage size in GB assuming 16-bit precision for uncompressed models and Golomb code-based compression. We find that on $\mathtt{28}$ of $\mathtt{32}$ experimental configurations \methodshort{} improves upon the performance of the original QLoRA models while compressing the LoRA module between $\mathtt{10x}-\mathtt{50x}$ in terms of storage costs. \methodshort{} leads to an improvement of $\mathtt{0.54}\%$, $\mathtt{1.06}\%$, $\mathtt{3.44}\%$, and $\mathtt{4.16}\%$ on MMLU for the $\mathtt{LLaMA}$ $\mathtt{7B}$, $\mathtt{13B}$, $\mathtt{33B}$, and $\mathtt{65B}$ parameter models, respectively. In Table~\ref{tab:llama2}, we observe similar compression and improvements for $\mathtt{LLaMA2-70B}$ model. To sum, \methodshort{} provides better results while also reducing the QLoRA module size. For example, on the $\mathtt{65B}$ $\mathtt{LLaMA}$ base model it reduces the storage size from $\mathtt{1.5}$GB to $\mathtt{110}$MB while improving the MMLU performance by a large margin of $4.16\%$.

\textbf{Discussion.} A few important conclusions about \methodshort{} can be derived from these results: (1) \methodshort{} can compress all QLoRA models by a factor of at least $\mathtt{10x}$. (2) Larger base models allow for more compressible LoRA modules. We get a compression factor of approximately $\mathtt{16x}$, $\mathtt{20x}$, $\mathtt{16x}$, and $\mathtt{26x}$ for $\mathtt{7B}$, $\mathtt{13B}$, $\mathtt{33B}$, and $\mathtt{65B}$ parameter models respectively. (3) A similar trend is found in performance -- the performance gap between the original and the compressed LoRA module increases with model size from $\mathtt{0.54}\%$ for the $\mathtt{7B}$ model to $\mathtt{4.16}\%$ for the $\mathtt{65B}$ model. If this scaling law continues, it means that the utility of methods like \methodshort{} will increase as models become larger and/or their zero-shot performance improves.

\subsection{Compressing Other PEFT Updates}
\label{sec:peft_small}

The finding that scaling the base model makes the PEFT modules more compressible and more performant brings up the question as to whether \methodshort{} is still effective at smaller scales. We perform experiments on two widely used PEFT methods, (IA)$^3$~\citep{liu2022tfew} and LoRA~\citep{hu2021lora}, with three models, T5-Base and T5-Large~\citep{colin2020exploring}, and T0-3B~\citep{sanh2022multitask}. Specifically, we compress (IA)$^3$ and LoRA modules trained on $\mathtt{7}$ classification tasks from the GLUE benchmark~\citep{wang2018glue} belonging to three categories: Natural Language Inference (MNLI~\citep{williams2018mnli}, RTE~\citep{bentivogli2009rte}, QNLI~\citep{rajpurkar2016squadqnli}, WNLI~\citep{levesque2012winogradwnli}), Sentiment Analysis (SST2~\citep{socher2013sst2}), and Paraphrase Detection (MRPC~\citep{dolan2005automaticallymrpc}, QQP~\citep{wang2018glue}). For hyperparameter selection ($\alpha, k$), we use the same grid and validation procedure as described in \S\ref{sec:peft_llama}.

\textbf{Outcomes.} In Table~\ref{tab:main_goodzs_peft_test}, we present the average performance on the $\mathtt{7}$ aforementioned GLUE tasks (per-dataset results are provided in Appendix~\ref{sec:app_ind_results}) along with the average checkpoint size in MB (in subscripts) for three base models with both (IA)$^3$ and LoRA adapters. We find that even with smaller base models, \methodshort{} compress the PEFT modules by a factor of $\mathtt{12x-25x}$ with minimal to no loss in performance. These results demonstrate that even at smaller scales \methodshort{} can lead to substantial compression. Additionally, we performed some experiments with BERT~\citep{devlin2018bert}, RoBERTa~\citep{liu2019roberta}, and T5v1.1~\citep{colin2020exploring} models that are not multitask-trained and/or have weak zero-shot performance (i.e.\ they generally require additional finetuning to perform well on any downstream tasks). We present the results for these models in Appendix~\ref{sec:app_bad_zs}, where we observe that compression works well for LoRA with minimal performance loss. However, for (IA)$^3$ we observe significant performance drops which suggest that zero-shot performance may be important to enable \methodshort{} compression of (IA)$^3$-based models.

\begin{figure}[t!]
  \centering
  \vspace{-0.7cm}
    \begin{minipage}{0.49\linewidth}
        \centering
        \captionsetup{type=table}
        \caption{\label{tab:main_goodzs_full_test} \textbf{\methodshort{} extends compression benefits to fully fine-tuned residuals.} Average test set performance (and storage size in GB) over 7 GLUE tasks for original and \methodshort{}-compressed fully fine-tuned model's task vectors. Across various model architectures and sizes, \methodshort{} compresses models by $\mathtt{12x-19x}$ with near-lossless performance, and even slight improvements for some models.}

        \resizebox{\linewidth}{!}{  
        \begin{tabular}{lccc}
        \toprule
        
        \textbf{$\mathtt{Model}$ ($\downarrow$)} &  $\mathtt{Original}$  & \methodshort{} & $\mathtt{Improvement}$ \\
         \midrule

        $\mathtt{BERT-Base}$ & 87.2$_{(0.21)}$ & 86.8$_{(0.011)}$ & \textcolor{brickred}{-0.4} / \textcolor{color2}{$\mathtt{19x}$}  \\
        $\mathtt{BERT-Large}$ & 86.3$_{(0.64)}$ & 86.1$_{(0.036)}$ & \textcolor{brickred}{-0.2} / \textcolor{color2}{$\mathtt{18x}$}  \\
        \midrule
        $\mathtt{RoBERTa-Base}$ & 85.5$_{(0.24)}$ & 83.3$_{(0.013)}$ & \textcolor{brickred}{-2.2} / \textcolor{color2}{$\mathtt{18x}$}  \\
        $\mathtt{RoBERTa-Large}$ & 88.6$_{(0.68)}$ & 89.2$_{(0.052)}$ & \textcolor{color2}{+0.6} / \textcolor{color2}{$\mathtt{13x}$}  \\
        \midrule
        $\mathtt{T5v.1-Base}$ & 74.1$_{(0.47)}$ & 75.8$_{(0.032)}$ & \textcolor{color2}{+1.7} / \textcolor{color2}{$\mathtt{15x}$}  \\
        $\mathtt{T5v.1-Large}$ & 84.0$_{(1.5)}$ & 82.2$_{(0.11)}$ & \textcolor{brickred}{-1.8} / \textcolor{color2}{$\mathtt{14x}$}  \\
        \midrule
        $\mathtt{T5-Base}$ & 82.8$_{(0.43)}$ & 78.1$_{(0.032)}$ & \textcolor{brickred}{-4.7} / \textcolor{color2}{$\mathtt{13x}$}  \\
        $\mathtt{T5-Large}$ & 85.2$_{(1.41)}$ & 84.7$_{(0.12)}$ & \textcolor{brickred}{-0.5} / \textcolor{color2}{$\mathtt{12x}$} \\
        
        \bottomrule
        \end{tabular}
        }
  \end{minipage}
  \hfill
  \begin{minipage}{0.49\linewidth}
    \centering
        \captionsetup{type=table}
        \caption{\label{tab:timing} \textbf{\methodshort{} enables order-of-magnitude reduction in model transmission and loading latency.} 
        We report the wall clock time (mean and standard deviation) for two practical scenarios: downloading LLaMA model checkpoints (7B-65B) from a simulated internet server to a local machine, and transferring checkpoints from CPU to GPU memory, comparing original and \methodshort{}-compressed versions. \methodshort{} reduces download times by up to $\mathtt{32x}$ and CPU-to-GPU loading times by up to $\mathtt{25x}$, highlighting the practical advantages in deployment and serving efficiency.}

        \resizebox{\linewidth}{!}{      
        \begin{tabular}{lcccc}
        \toprule
        \textbf{Model ($\downarrow$)} & \multicolumn{2}{c}{\textbf{Internet $\rightarrow$ Local (seconds)}} & \multicolumn{2}{c}{\textbf{CPU $\rightarrow$ GPU (milliseconds)}} \\
        \cmidrule(lr){2-3} \cmidrule(lr){4-5} 
        & \textbf{Original} & \methodshortbold{} & \textbf{Original} & \methodshortbold{} \\
        \midrule
        
        $\mathtt{LLaMA-7B}$  & $11.21_{2.44}$ & $1.16_{0.04}$ & $134.28_{4.76}$ & $11.23_{5.22}$ \\
        $\mathtt{LLaMA-13B}$ & $16.85_{3.83}$ & $1.75_{0.30}$ & $186.60_{5.42}$ & $23.09_{0.78}$ \\
        $\mathtt{LLaMA-33B}$ & $32.31_{6.76}$ & $2.46_{0.12}$ & $307.29_{55.59}$ & $18.00_{4.34}$ \\
        $\mathtt{LLaMA-65B}$ & $83.17_{9.14}$ & $2.59_{0.14}$ & $475.26_{66.51}$ & $18.60_{5.67}$ \\ 
        \bottomrule
        \end{tabular}
        }    
  \end{minipage}
\end{figure}

\subsection{Compressing Fully Fine-tuned Models}
\label{sec:full_compress}

\textbf{Experimental Setup.}  To assess the broader applicability of \methodshort{}, we investigate its effectiveness beyond PEFT modules and explore its ability to compress task vectors produced by full-model fine-tuning. We adopt the experimental setting from \S~\ref{sec:peft_small} to fine-tune the $\mathtt{7}$ classification tasks from the GLUE benchmark using full fine-tuning, and then compress the resulting fully fine-tuned task vectors using \methodshort{}. We evaluate our method on four different model architectures -- BERT~\citep{devlin2018bert}, RoBERTa~\citep{liu2019roberta}, T5-v1.1~\citep{colin2020exploring}, and T5~\citep{colin2020exploring} -- across two model sizes (Base and Large) for each architecture.

\textbf{Outcomes.} Table~\ref{tab:main_goodzs_full_test} presents the average test set performance over the $\mathtt{7}$ GLUE tasks for both original and \methodshort{}-compressed fully fine-tuned modelsWe observe that \methodshort{} effectively compresses fully fine-tuned models, achieving $\mathtt{12x-19x}$ compression ratios with minimal performance degradation. Notably, for T5v1.1-base and RoBERTa-large models, \methodshort{} even leads to performance improvements of $\mathtt{1.7}\%$ and $\mathtt{0.6}\%$ respectively, while simultaneously reducing model size by $\mathtt{15x}$ and $\mathtt{13x}$.

\textbf{Discussion.} These results demonstrate that \methodshort{} is not limited to compressing parameter-efficient modules but can also be effectively applied to compress fully fine-tuned models. This broader applicability expands the potential use cases of \methodshort{} and highlights its versatility as a general model compression technique. The observed performance improvements in some cases, even with full fine-tuning compression, further suggest that \methodshort{} may act as a regularizer, potentially improving generalization.

\subsection{Reduced Transmission Cost and Loading Latency}
\label{sec:timing_latency}

\textbf{Experimental Setup.}  A key practical advantage of model compression is reduced storage and transmission costs. To quantify these benefits for \methodshort{}, we measure the wall clock time for two representative scenarios: (1) downloading a model checkpoint from a simulated internet server to a local machine, and (2) loading a model checkpoint from CPU memory to GPU memory. We perform these measurements for both original QLoRA checkpoints and their \methodshort{}-compressed counterparts for LLaMA models of sizes $\mathtt{7B}$, $\mathtt{13B}$, $\mathtt{33B}$, and $\mathtt{65B}$For each scenario and model configuration, we repeat the measurement $\mathtt{10}$ times and report the mean and standard deviation of the wall clock times.

\textbf{Outcomes.} Table~\ref{tab:timing} presents the results for transmission latency and loading timeAs expected, \methodshort{}-compressed checkpoints exhibit significantly reduced transmission and loading times compared to the original checkpoints across all model sizes and both scenariosFor example, downloading the $\mathtt{LLaMA-65B}$ \methodshort{} checkpoint from the internet is approximately $\mathtt{32x}$ faster than downloading the original checkpoint. Similarly, loading the \methodshort{}-compressed $\mathtt{LLaMA-65B}$ checkpoint from CPU to GPU is about $\mathtt{25x}$ faster.

\textbf{Discussion.}  These measurements showcase \methodshort{}'s substantial practical value beyond their size. The order-of-magnitude speedups in download and loading/offloading times significantly accelerate model deployment and improve multi-expert serving efficiency, especially in dynamic or resource-limited settings\methodshort{} thus offers key real-world advantages for efficient model utilization.

\subsection{\texorpdfstring{\methodshort{}}{methodshort} vs. Other PEFT Methods}
\label{sec:pareto_peft}

\begin{wrapfigure}[21]{r}{0.5\textwidth}
    \centering
    \includegraphics[width=1.0\linewidth]{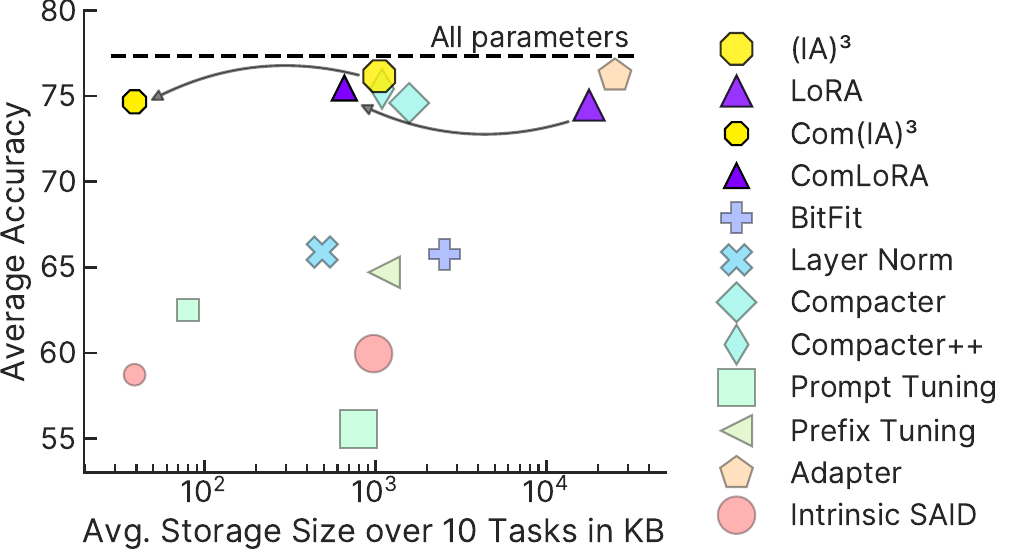}
    % \vspace{-0.2cm}
    \caption{\label{fig:pareto_peft} \textbf{\methodshort{} are Pareto-optimal.} Performance vs storage size for multiple PEFT methods averaged over $\mathtt{11}$ tasks. A PEFT method is Pareto-optimal if it attains better performance (higher on the y-axis) than all methods that use less storage space (to the left on the x-axis). In particular, \methodia{} performance is comparable to PEFT methods that require $\mathtt{1000\times}$ more storage space.}
\end{wrapfigure}

\textbf{Experimental Setup.} Next, we compare the (IA)$^3$ and LoRA checkpoints compressed by \methodshort{} with various other PEFT methods to determine whether \methodshort{} produces Pareto-optimal parameter-efficient fine-tuning in terms of Storage Size and Performance. For this, we use the T0-3B~\citep{sanh2022multitask} model and train a wide range of PEFT methods on the $\mathtt{11}$ held-out datasets from \citet{sanh2022multitask} -- specifically, sentence completion (COPA \citep{copa}, H-SWAG \citep{zellers2019hellaswag}, and Story Cloze \citep{sharma2018tackling} datasets), natural language inference (three splits of ANLI \citep{nie2019adversarial}, CB \citep{cb}, and RTE \citep{dagan2005pascal}), coreference resolution (WSC \citep{wsc} and Winogrande \citep{sakaguchi2020winogrande}), and word sense disambiguation (WiC \citep{pilehvar2018wic}). For each task, from the training set, we select $\mathtt{200}$ example for the validation set and then use the first template from Prompt Source~\citep{bach2022promptsource} both during training and evaluation. We perform experiments with $\mathtt{10}$ different PEFT methods from \citet{liu2022tfew} -- LoRA~\citep{hu2021lora}, (IA)$^3$~\citep{liu2022tfew}, BitFit~\citep{zaken2021bitfit}, LayerNorm, Adapters~\citep{houlsby2019parameter_adapter}, Compacter and Compactor++~\citep{karimi2021compacter}, Prompt Tuning~\citep{lester2021prompttuning}, Prefix Tuning~\citep{li2021prefixtuning}, and Intrinsic SAID~\citep{aghajanyan2020intrinsic}.

\textbf{Outcomes.} In Figure~\ref{fig:pareto_peft}, we plot the average performance over the $\mathtt{11}$ tasks and the checkpoint sizes in KB for $\mathtt{10}$ PEFT Method and when using \methodshort{} on LoRA and (IA)$^3$ checkpoints, i.e. \methodlora{} and \methodia{}. We find that \methodshort{} reduces the storage size for both LoRA and (IA)$^3$ by more than an order of magnitude with minimal reduction in performance. From this plot, \methodshort{} is Pareto-optimal, i.e. for any given storage budget, \methodshort{} outperforms all other PEFT methods. Notably, \methodia{} exhibits only a minor performance degradation compared to full-model fine-tuning while being one of the most space-efficient PEFT methods. Lastly, we note that for \methodshort{} you can trade-off performance for storage cost by varying the density $k$ to obtain models of different sizes. Hence, \methodia{} and \methodlora{} could be made even more space efficient.

\textbf{Discussion.} Figure~\ref{fig:pareto_peft} positions \methodshort{} as a highly competitive PEFT technique, not just a compression methodWhile methods like Prompt Tuning and Prefix Tuning offer parameter efficiency, \methodshort{}, especially \methodia{}, achieves a significantly better balance of performance and size. Traditional full fine-tuning, while offering slightly higher peak performance, is orders of magnitude larger in sizeThis Pareto-optimality highlights the practical advantages of \methodshort{} in resource-constrained scenarios. Furthermore, while we focus on Pareto optimality against other PEFT methods here, \methodshort{} also provides a strong compression baseline for any PEFT technique; applying \methodshort{} to other PEFT outputs could further enhance their storage efficiency.

\begin{figure}[t!]
  \centering
    \begin{minipage}{0.39\linewidth}
        \captionsetup{type=figure}
        \includegraphics[width=0.85\linewidth]{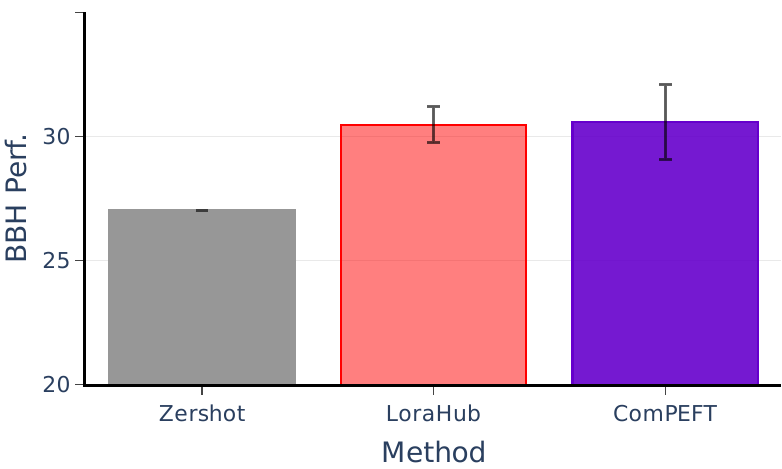}
        \caption{\label{fig:lorahub} \textbf{\methodshort{} facilitates compositional generalization.} Average performance of LoraHub and \methodshort{} for compositional generalization on BBH.}
  \end{minipage}
  \hfill
  \begin{minipage}{0.59\linewidth}
    \centering
    \captionsetup{type=table}
    \caption{\label{tab:merging} \textbf{\methodshort{} compressed checkpoints lead to better merged models.} Average test set results on $\mathtt{7}$ GLUE tasks when employing different merging methods on the uncompressed checkpoints and compressed \methodshort{} checkpoints.}
    
    \resizebox{\linewidth}{!}{  
    \begin{tabular}{lcccccccc}
    \toprule
    
    \textbf{Method ($\downarrow$)} & \multicolumn{2}{c}{$\mathtt{\textbf{T5-Base}}$}  & \multicolumn{2}{c}{$\mathtt{\textbf{T5-Large}}$} & \multicolumn{2}{c}{$\mathtt{\textbf{T0-3B}}$} \\
     \cmidrule(lr){2-3} \cmidrule(lr){4-5} \cmidrule(lr){6-7}
    
     & \textbf{(IA)}$^\textbf{3}$ & \textbf{LoRA} & \textbf{(IA)}$^\textbf{3}$ & \textbf{LoRA} & \textbf{(IA)}$^\textbf{3}$ & \textbf{LoRA} \\
     \midrule
    
    \textbf{Averaging} &  53.7 & 49.3 & 55.4 & 50.2 & 74.5 & 73.1  \\
    \midrule
    \textbf{Task Arithmetic~(TA)} & \textbf{60.0} & 52.8 & \textbf{62.7} & 61.6 & 77.8 & 75.0 \\
    \methodshortbold{} \textbf{~+ ~TA}  & 59.7 & \textbf{53.9} & 61.9 & \textbf{64.9} & \textbf{80.0} & \textbf{75.6} \\ 
    \midrule
    \textbf{TIES-Merging} & 55.5 & 49.2 & \textbf{61.3} & 57.3 & 71.7 & 73.4  \\
    \methodshortbold{} \textbf{~+ ~TIES} & \textbf{55.6} & \textbf{49.2} & 60.4 & \textbf{61.4} & \textbf{76.2} & \textbf{75.8}  \\
    
    \bottomrule
    \end{tabular}
    }
  \end{minipage}
\end{figure}

\subsection{Cross-Task Generalization via Dynamic LoRA Module Composition}
\label{sec:lorahub}

\textbf{Experimental Setup.}
As highlighted in \S\ref{sec:introduction}, a key motivation is to enable efficient serving of numerous experts, particularly in scenarios requiring dynamic adaptation to novel tasks. Cross-task generalization, using compositional methods like LoraHub which dynamically select, load, and compose expert modules, exemplifies such a scenario where communication bottlenecks are critical. Therefore, to assess \methodshort{}'s ability to facilitate efficient expert module serving in this demanding downstream application, we examine its impact on the composability of the resulting PEFT modules for cross-task generalization. Given a set of expert models and an unseen downstream task with few training examples, the goal is to combine a subset of these expert modules to attain a model that performs well on the unseen task.

For this, we follow the LoraHub~\citep{huang2023lorahub} method and their experimental setting. We use the Flan-T5-large~\citep{chung2022scaling_flant5} model as it exhibits strong zero-shot and few-shot capabilities. We consider nearly $200$ distinct (tasks,  instruction) pairs that were utilized to train the Flan-T5 model and use the LoRA modules trained on these tasks as expert\footnote{\href{https://hf.co/models?search=lorahub}{hf.co/models?search=lorahub}}. Following \citet{huang2023lorahub}, when learning a new unseen task, we randomly select $\mathtt{N}$ LoRA modules denoted by $\{L_i = (A_i, B_i)\}_{i=1}^N$ and compose them as
\begin{equation}
   L_m = A_m B_m = \left( \sum_{i=1}^N w_iA_i \right) \left( \sum_{i=1}^N w_iB_i \right),
   \label{eqn:lorahub}
\end{equation}
where $A_m$, $B_m$ are the matrics of the composed LoRA module and $w_i$ are parameters that are learned on the few-shot examples from the unseen tasks using the gradient-free Shiwa optimizer~\citep{liu2020versatile_opt}. Following LoraHub, we use $\mathtt{N} =20$ and treat the $\mathtt{27}$ diverse tasks from the Big-Bench Hard (BBH) benchmark~\citep{suzgun2022challenging_bbh} as our unseen evaluation tasks. All the tasks are multiple-choice questions and we employ Exact Match (EM) as our evaluation metric. Error bars in Figure~\ref{fig:lorahub} represent standard deviation across the BBH tasks performance.

\textbf{Outcomes.} In Figure~\ref{fig:lorahub}, we report the average performance and standard deviation (over 5 seeds) when using the LoraHub method on the original checkpoints and the \methodshort{}-compressed checkpoints. We find that the \methodshort{}-compressed checkpoints exhibit similar compositional abilities as the original uncompressed checkpoints. This is a crucial finding: even with extreme compression, the modules retain the necessary properties for effective cross-task composition. Hence, \methodshort{} checkpoints can be communicated quickly over high latency networks for dynamic module swapping, while maintaining their compositional abilities.

\textbf{Discussion.} The preservation of compositional generalization performance after \methodshort{} compression is a significant result. It directly addresses the practical challenge of serving numerous expert modules for complex tasksBy drastically reducing module size, \methodshort{} makes dynamic module composition via methods like LoraHub far more efficient and scalable, enabling faster download times and reduced memory footprint during run-time module swappingThis experiment validates that \methodshort{} is not only a compression technique but also a facilitator for advanced applications requiring efficient expert module management and communication.

\subsection{Merging Compressed PEFT Modules}
\label{sec:merging}

\textbf{Experimental Setup.} Next, we examine the effectiveness of \methodshort{} when merging models~\citep{choshen2022fusing,matena2021merging} by comparing the merging of compressed or uncompressed models. We follow the experimental setting (including base models, PEFT methods, and datasets) from the previous section and merge the $\mathtt{7}$ GLUE tasks to produce a multitask model.
We then report the average performance of the merged across all tasks. 
We use two methods to merge task vectors, namely, Task Arithmetic~\citep{ilharco2023editing} and TIES-Merging~\citep{yadav2023ties-merging}. We used the code from the original authors for both merging methods.

\textbf{Outcomes.} As demonstrated in Table~\ref{tab:merging}, in $\mathtt{9}$ out of $\mathtt{12}$ scenarios, the \methodshort{} checkpoints lead to better merged models compared to the original checkpoints, with the notable exception of (IA)$^3$ on T5 models. Notably, in stronger models like T0-3B, \methodshort{}-compressed checkpoints not only reduce the size by approximately $\mathtt{15x}$ but also improve the merged model's performance by $\mathtt{2.4}\%$ on average. One possible explanation is that \methodshort{} acts as a regularizer, removing less important parameter updates and potentially leading to smoother, more generalizable loss landscape that merge more effectively. This shows \methodshort{}'s efficacy in both minimizing storage and communication overheads and improving the model merging performance.

\section{Additional Results and Analysis}
\label{sec:additional_results}

\begin{figure*}[t!]
    \vspace{-1cm}
  \centering
  \begin{subfigure}[b]{0.19\textwidth}
        \centering
        \includegraphics[width=\linewidth]{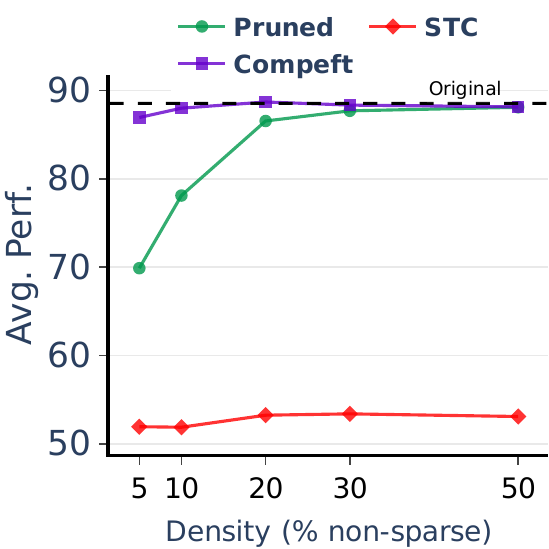}
        \caption{T0-3B}
  \end{subfigure}
  \hfill
  \begin{subfigure}[b]{0.19\textwidth}
        \centering
        \includegraphics[width=\linewidth]{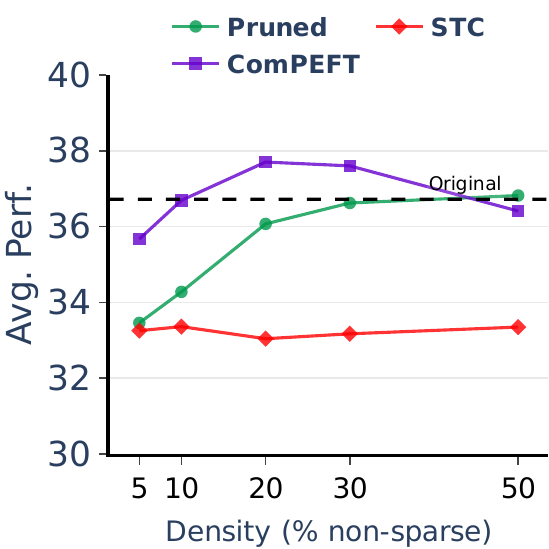}
        \caption{LLaMA-7B}
  \end{subfigure}
  \hfill
  \begin{subfigure}[b]{0.19\textwidth}
        \centering
        \includegraphics[width=\linewidth]{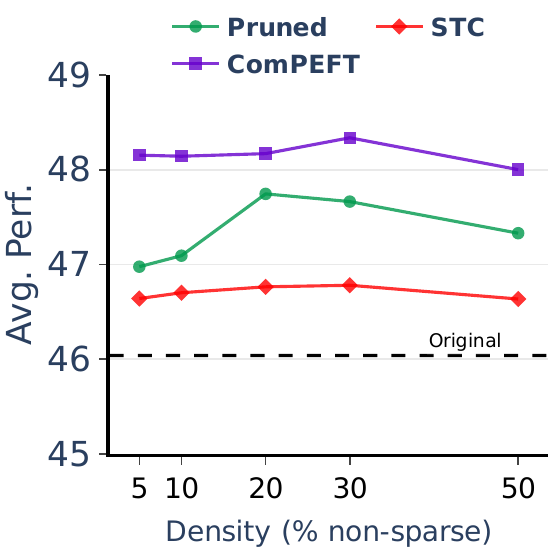}
        \caption{LLaMA-13B}
  \end{subfigure}
  \hfill
  \begin{subfigure}[b]{0.19\linewidth}
        \centering
        \includegraphics[width=\linewidth]{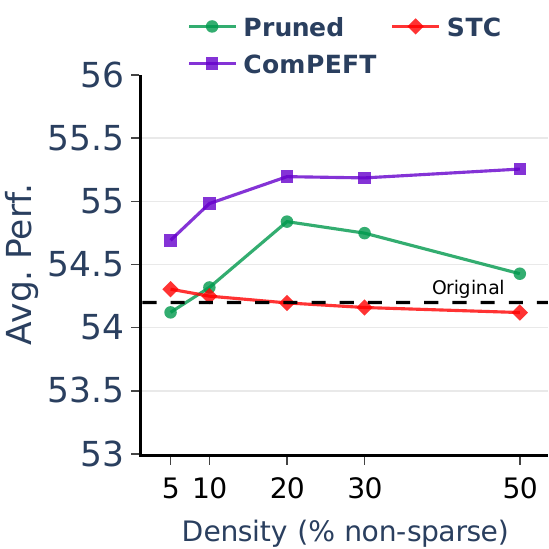}
        \caption{LLaMA-33B}
  \end{subfigure}
  \hfill
  \begin{subfigure}[b]{0.19\linewidth}
        \centering
        \includegraphics[width=\linewidth]{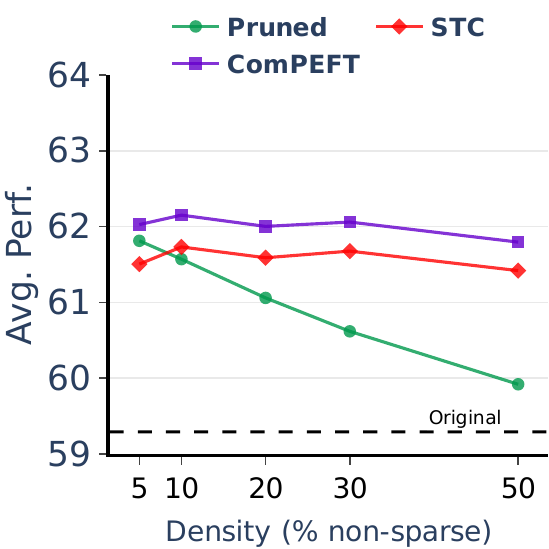}
        \caption{LLaMA-65B}
  \end{subfigure}

    \vspace{-0.2cm}
  \caption{\label{fig:ablations} \textbf{\methodshort{} outperforms STC and all method steps are crucial.} Average performance as density $\mathtt{k}$ of the compressed checkpoint increases. We show results for compressing LoRA modules trained over different models with sizes ranging from $\mathtt{3B-65B}$ and compare them with baselines and ablate method components.}
  \vspace{-0.3cm}
\end{figure*}

\subsection{Ablation of \texorpdfstring{\methodshort{}}{methodshort}  Components}

\textbf{Experimental Setup.} To understand the contribution of the individual steps of \methodshort{}, we now perform a brief ablation study. In \methodshort{} there are two main steps: (1) Sparsifying the direction vectors, and (2) Quantizing the magnitudes to a scalar with scaling factor $\alpha$. Hence, we compare with two ablated versions: \textit{Pruned} (only sparsification, magnitudes reset to zero, no quantization, no scaling), and Sparse Ternary Compression (\textit{STC})~\citep{sattler2019robust_stc} (ternary quantization with mean magnitude scaling, no tuned $\alpha$). We also include the uncompressed \textit{original} model as a baseline. We provide these ablations for the experimental settings from \S~\ref{sec:peft_llama} and \ref{sec:pareto_peft} where the model sizes range from $\mathtt{3B-65B}$.

\textbf{Outcomes.} In Figure~\ref{fig:ablations}, we plot the average validation set performance over tasks as a function of the density ($k$) of the compressed model. From these results, we make a few observations: (1) \methodshort{} almost always performs better than both STC and the Pruned version for all model sizes and sparsity levels. (2) \methodshort{} almost always performs better than or similar to the original model's performance for all sparsity levels. In contrast, for smaller model sizes of $\mathtt{3B}$ and $\mathtt{7B}$, STC's performance is much worse than the original models. This highlights the importance of the scaling $\alpha$ as proposed in \methodshort{}, which allows us to recover the performance lost due to pruning and ternary compression without computationally expensive retraining. (3) At low density, the performance of \textit{Pruned} is much worse than \methodshort{} and this gap reduces as the density increases. However, note that the size of \methodshort{} is much smaller than the \textit{Pruned} baseline due to ternarization. (4) At larger base model sizes ($\geq\mathtt{13B}$), all the methods at all density levels perform similarly to or better than the original LoRA checkpoint, suggesting increased robustness to compression choices at scale.

\textbf{Discussion.} The ablation study clearly demonstrates the contribution of each component. The superior performance of \methodshort{} over the \textit{Pruned} variant underscores the importance of ternary quantization and scalar scaling in maintaining performance after sparsification. The advantage over STC highlights the benefit of tuning the scaling factor $\alpha$ rather than using a fixed magnitude scaling like mean magnitude. These results validate our design choices and show the utility of sparsification, ternarization, and tuned scalar scaling.

\begin{figure*}[t!]
  \centering
  \begin{subfigure}[b]{0.19\textwidth}
        \centering
        \includegraphics[width=\linewidth]{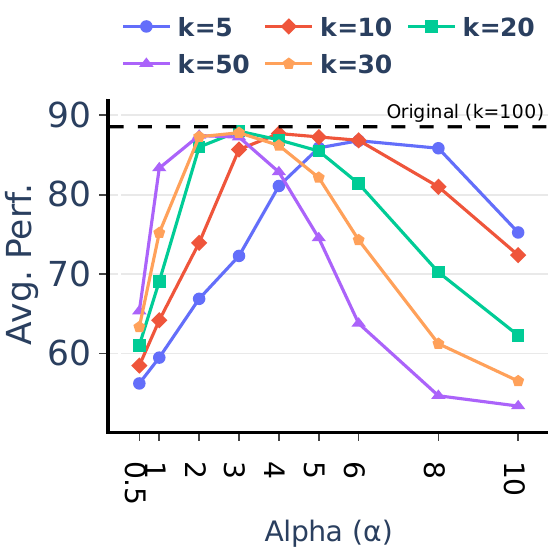}
        \caption{T0-3B}
  \end{subfigure}
  \hfill
  \begin{subfigure}[b]{0.19\textwidth}
        \centering
        \includegraphics[width=\linewidth]{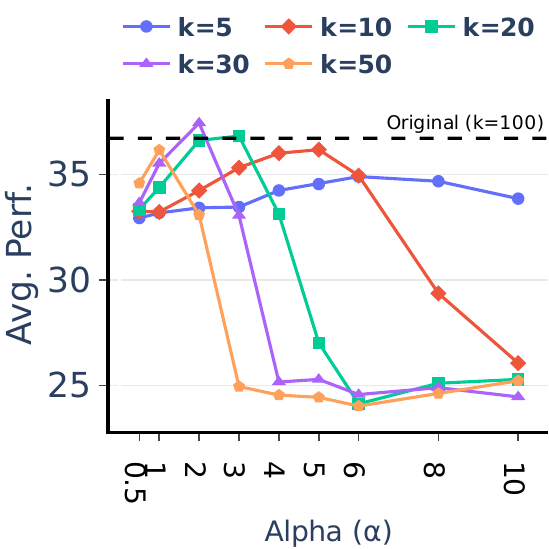}
        \caption{LLaMA-7B}
  \end{subfigure}
  \hfill
  \begin{subfigure}[b]{0.19\textwidth}
        \centering
        \includegraphics[width=\linewidth]{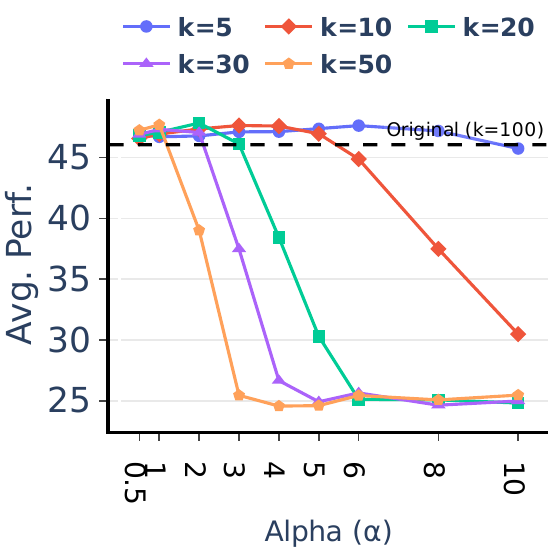}
        \caption{LLaMA-13B}
  \end{subfigure}
  \hfill
  \begin{subfigure}[b]{0.19\linewidth}
        \centering
        \includegraphics[width=\linewidth]{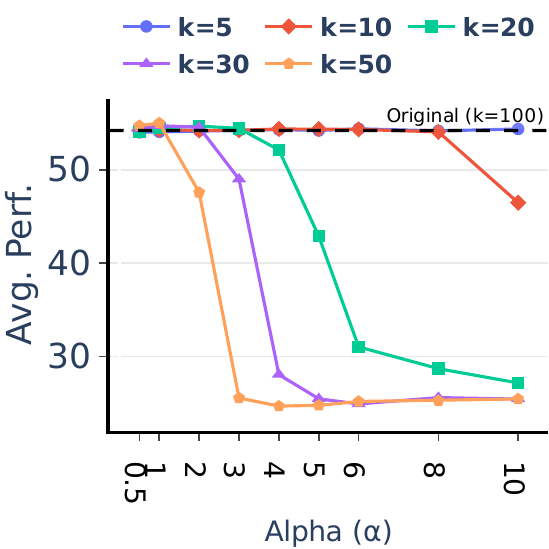}
        \caption{LLaMA-33B}
  \end{subfigure}
  \hfill
  \begin{subfigure}[b]{0.19\linewidth}
        \centering
        \includegraphics[width=\linewidth]{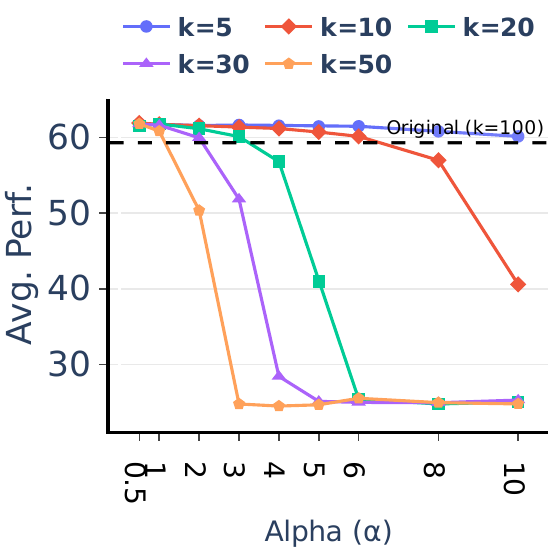}
        \caption{LLaMA-65B}
  \end{subfigure}

    \vspace{-0.2cm}
  \caption{\label{fig:sparsity_scaling} \textbf{Larger models do not require explicit tuning of $\alpha$.} Performance vs $\alpha$ for various denisty levels for \methodshort{}.}
  \vspace{-0.5cm}
\end{figure*}

\subsection{Effect of Sparsity and Scaling on \texorpdfstring{\methodshort{}}{methodshort} }
\label{sec:spar_scale}

\textbf{Experimental Setup.} For \methodshort{}, we analyze the effect of different levels of sparsity and the scaling value $\alpha$ on the performance of the compressed checkpoints. We present this analysis for $\mathtt{T0-3B}$ and $\mathtt{LLaMA}$ as the base models; the experimental settings are similar to \S~\ref{sec:peft_llama} and \ref{sec:pareto_peft} where the model sizes range from $\mathtt{3B-65B}$. We provide results for different values of the density $k$ (sparsity = $100-k$), specifically, the values $k \in \{5, 10, 20, 30, 50\}$ and different values of $\alpha \in \{0.5, 1, 2, 3, 4, 5, 6, 8, 10\}$. 

\textbf{Outcomes.} In Figure~\ref{fig:sparsity_scaling}, we plot the average validation set performance across all tasks with respect to the scaling coefficient $\alpha$. We make the following observations; (1) For smaller base-model sizes ($\mathtt{3B}$ and $\mathtt{7B}$) and across density values, we find a similar trend -- as the value of $\alpha$ increases, the average validation performance first increases and then drops. (2) As the value of $k$ increases, the optimal value of $\alpha$ tends to becomes smaller. For example, for the T0-3B base model, the optimal value $\alpha$ for $k=50$ is between $\mathtt{2-3}$ while for $k=5$ the optimal $\alpha$ is in the range $\mathtt{5-8}$. 
(3) For bigger base-models ($\mathtt{\geq 13B}$) and low density ($k \in \{5,10,20\}$) the variation in performance as $\alpha$ changes is smaller. (4) Lastly, as the base-model size increases, smaller values of $\alpha \in (0.5, 2)$ and a bigger range of values start to work better. Hence, for large models, the need for tuning $\alpha$ can be removed. For models with $\mathtt{\geq 13B}$ parameters and high sparsity $k \leq \mathtt{20}$, we recommend setting $\alpha=1$.

\textbf{Discussion.} This analysis highlights the interplay between sparsity and scaling in ComPEFT. For smaller models, fine-tuning $\alpha$ is important to maximize performance at a given sparsity level. However, for larger models, the method becomes more robust, and a fixed scaling factor (like $\alpha=1$) can be sufficient, especially at higher sparsity. This robustness for larger models simplifies the application of ComPEFT in practice, as it reduces the need for extensive hyperparameter search and makes the method more readily deployable for large language models where computational efficiency is paramount.

%% file: sections/related_work.tex
\label{sec:related_work}
\vspace{-2pt}
\section{Related Work}

\textbf{Paremeter Efficient Fine-Tuning.} 
Several parameter-efficient techniques~\citep{lester2021prompttuning,li2021prefixtuning,houlsby2019parameter_adapter,zaken2021bitfit} have emerged as efficient alternatives to full fine-tuning in the field of pre-trained language models (PLMs). These methods introduce a small number of additional parameters to PLMs~\citep{colin2020exploring,touvron2023llama} enabling efficient fine-tuning. LoRA~\citep{hu2021lora} incorporates trainable low-rank matrices into transformer layers. In Contrast, (IA)$^3$~\citep{liu2022tfew} learns a new set of parameters to rescale the model activations. Recently, QLoRA~\citep{dettmers2023qlora} proposed training LoRA modules over a $\mathtt{4}$-bit quantized base model to further save the memory. 

\textbf{Network Pruning and Federated Learning.} Neural network pruning techniques have garnered attention for reducing computational costs~\citep{DBLP:journals/corr/abs-1710-09282,DBLP:journals/ijon/LiangGWSZ21} by removing redundant parameters while preserving performance~\citep{DBLP:conf/iclr/ZhuG18,DBLP:conf/iclr/LiuSZHD19,DBLP:conf/iclr/FrankleC19,DBLP:journals/corr/abs-1902-09574,DBLP:conf/acl/XiaZC22}. Among these, magnitude-based pruning~\citep{han2015learning,DBLP:conf/ijcai/LiQJLT18,DBLP:conf/iclr/LeePMAS21} selects parameters based on magnitudes. Pruning is valuable in federated learning due to high communication costs over slow networks. Atomo~\citep{wang2018atomo} minimizes gradient variance through unbiased sparsification, while QSGD~\citep{alistarh2017qsgd} offers a communication-convergence trade-off by quantizing gradients. SignSGD~\citep{bernstein2018signsgd} further converts gradients to binary sign vectors. TernGrad~\citep{wen2017terngrad} and STC~\citep{sattler2019robust_stc} combine sparsification and quantization.

\textbf{Model Merging and Compositional Generalization.}
Various merging methods~\citep{ortizjimenez2023tangent,wortsman2021robust,wortsman2022model,ilharco2022patching,rame2022modelrat,yu2023mario_merging} aim to combine fine-tuned models for improved performance in various applications. \citet{choshen2022fusing} performs direct averaging of the model weights while Task Arithmetic~\citep{ilharco2023editing} generates task vectors and performs arithmetic operations to create multitask checkpoints. 
\citet{ortizjimenez2023tangent} offer theoretical insights into model merging by using the weight disentanglement property. TIES-Merging~\citep{yadav2023ties-merging} identifies the issue of parameter interference in model merging and tackles it by trimming low-magnitude parameters, resolving sign disagreements, and disjointly merging parameters with consistent signs. \citet{ponti2023combining} performed CG by jointly learning adapters and a routing function to allocate skills to tasks, while~\citet{caccia2023multi_mhr} analyzes task routing for more efficient cross-task generalization. LoraHub~\citep{huang2023lorahub} employs gradient-free optimization to retrieve and merge expert modules for unseen tasks while \citet{muqeeth2024phatgoose} focus on zero shot compositional generalization. 
\citet{pfeiffer2023modular} provides an overview of PEFT methods, model merging, and compositional generalization methods.

% \newtext
% \textbf{Parameter Updates Compression.}
% \color{black}

% Recent advancements in the field of multi-task learning have aimed to improve generalization across various tasks. FLAN~\citep{wei2021finetuned}, T0~\citep{sanh2021multitask}, and InstructGPT~\citep{ouyang2022instructgpt} focus on enhancing the generalization capabilities of multi-task models. The CrossFit~\citep{ye2021crossfit} framework requires minimal labeled data for few-shot fine-tuning but relies on task names as hard prefixes, limiting generalization. ReCross~\citep{lin2022unsupervised_recross} reduces the need for labeled examples through retrieval but still involves a fine-tuning process. 

%% file: sections/appendix.tex
\section{Limitations}
\label{sec:limitation}

While \methodshort{} demonstrates significant promise, it is important to consider several limitations of this work.  Firstly, while average performance is strong, certain task types might exhibit reduced effectiveness or require specific hyperparameter tuning.  Secondly, we observed some performance sensitivity with \methodshort{} applied to (IA)$^3$ modules, especially on base models with weaker zero-shot capabilities, warranting further investigation into the interplay between base model properties, PEFT methods, and compression.  From a practical standpoint, while hyperparameter tuning for the scaling factor $\alpha$ becomes less crucial for larger models, it remains relevant for smaller models, adding a hyperparameter selection step to their deployment.  Future work could explore automated or adaptive methods to address this.  Furthermore, a rigorous theoretical understanding of \methodshort{} is still lacking. We do not have a definitive explanation for the observed performance improvements in some cases, nor why these improvements scale with model size, although noise reduction is a possible factor suggested by related works.  A deeper understanding of how fine-tuning updates encode information and how \methodshort{} interacts with this information is necessary for developing even more refined compression techniques.  Finally, to fully unlock the potential wall-clock speedups promised by \methodshort{}'s ternary vector representations, dedicated engineering effort is needed to develop custom Triton/CUDA kernels optimized for operations on sparse ternary data structures, as briefly discussed in our methodology section. These areas represent important avenues for future research, particularly in the context of efficiently serving and composing large numbers of expert PEFT modules for advanced applications.

\section{Implementation Details}
\label{sec:imp_details}

\subsection{Training Details}
\label{sec:app_training_details}

In our research, we utilized the following models, BERT-base, BERT-Large, RoBERTa-base, RoBERTa-large, T5v1.1-base, T5v1.1-large, T5-base, T5-large, Flan-T5-large, T0-3B, LLaMA $\mathtt{7B, 13B, 33B, 65B}$ models. The Flan-T5-Large and LLaMA models were not trained by us and were used by the authors of QLoRA~\citep{dettmers2023qlora} and LoraHub~\citep{huang2023lorahub}. For the experiments in \S\ref{sec:peft_small} and \S\ref{sec:full_compress} on the 7 GLUE~\citep{wang2018glue} tasks, we trained the large datasets (mnli, qnli, sst2, qqp) for 1 epoch and the small datasets (rte, mrpc, wnli) for 10 epochs. Whereas for the experiment in \S\ref{sec:pareto_peft}, we followed most of the hyperparameter configuration from the (IA)$^3$~\citep{liu2022tfew} paper and trained for 2500 steps with a batch size of 8. For each of the 11 datasets in \S\ref{sec:pareto_peft}, we selected 200 examples from the training set to be used as the validation set for best model selection as well as selecting the hyperparameters for \methodshort{}. Across all experiments to obtain the trained models we selected different learning rates for each dataset and PEFT method. For training (IA)$^3$ models we selected the learning rate from $\{1e-2, 1e-3, 1e-4, 1e-5\}$, for LoRA from $\{5e-2, 5e-3, 5e-4, 5e-5\}$, and for full model finetuning from $\{5e-3, 5e-4, 5e-5, 5e-6\}$. During the training process, bfloat16 was adopted to curtail GPU memory expenditure. For the purpose of evaluation, models from the T5 and T0 families were evaluated using rank classification to select the correct label. In this method, the model's log probabilities for all potential label strings are ranked. The model's prediction is deemed accurate if the choice ranked highest aligns with the correct answer. It should be noted that rank classification evaluation can accommodate both classification tasks and multiple-choice tasks.

\subsection{Compute Resources Used and Runtimes}
\label{sec:app_compute}
We executed all our experiments on Nvidia A6000 GPUs equipped with 48GB RAM. Training (IA)$^3$ and LoRA models on the T0-3B model for a single (\S\ref{sec:peft_small}, \S\ref{sec:pareto_peft}, and \S\ref{sec:merging}) task takes about 30 minutes to 4 hours depending on the dataset. For T5-Base and T5-Large models (\S\ref{sec:peft_small}, \S\ref{sec:merging}), based on dataset size, needed between 15 minutes and 2 hours per task. Experiments with QLoRA on LLaMA models were done using the original checkpoints from QLoRA paper~\citep{dettmers2023qlora} for all the 8 instruction tuning datasets and are supplied the authors of QLoRA here.\footnote{\href{https://huggingface.co/timdettmers?search\_models=qlora}{https://huggingface.co/timdettmers?search\_models=qlora}} 
The \methodshort{} compression experiments were efficient, with evaluations consuming between 10-30 seconds for the T5-Base, T5-Large, and T0-3B models. For LLaMA models, following QLoRA~\citep{dettmers2023qlora}, the hyperparameter selection is done on a small held-out subset of MMLU~\citep{hendrycks2020measuring_mmlu} benchmark and takes about 8 minutes, 14 minutes, 28 minutes, and 49 minutes for LLaMA $\mathtt{7B}$, $\mathtt{13B}$, $\mathtt{33B}$, and $\mathtt{65B}$ models respectively.

\subsection{Employed Datasets and Associated Licences}
\label{sec:license}
We use the following datasets in the paper with the following licenses. \\
Apache License 2.0: Flan V2, Self-Instruct, Chip2 \\
cc-by-nc-4.0: Alpaca \\
MIT License: Guanaco, Unatural Instructions, HH-RLHF, Longform \\
Not Found: GLUE

\subsection{Gradient Noise}\label{ap:gradient_noise}
Gradients are (almost) never 0 for any parameter, as all parameters somehow affect the result. Thus, we presume most updates in fine-tuning are not more than just noise, rather than learned updates. We compute the mean and standard deviation of the task vector of a LoRA model finetuned on LLaMa~\citep{touvron2023llama} base model and compare it with the base model. We find the mean of both the LoRA task vector and the base model is close to zero, however, the LoRA task vector has a small standard deviation of $\mathtt{0.0007}$ as compared to $\mathtt{0.0228}$ for the LLaMA base. This further confirms the hypothesis that most parameters are changed very little during fine-tuning.

\begin{table}[t!]
\centering

\caption{\label{tab:statistics} Statistics of the distribution of the task vectors for different model sizes and datasets.}
\begin{tabular}{llcccc}
\toprule
\textbf{Model ($\downarrow$)}   & \textbf{Dataset ($\downarrow$)}    & $\mathtt{TV_{mean}}$    & $\mathtt{TV_{std}}$ & $\mathtt{TV_{max}}$   & $\mathtt{TV_{min}}$   \\ 
\midrule
\multirow{2}{*}{$\mathtt{T0-3B}$}        & \textbf{storycloze} & 1.61E-02  & 0.1347 & 31.9909 & -4.8564 \\
& \textbf{winogrande} & 1.61E-02  & 0.1357 & 32.3445 & -4.7301 \\
\midrule
\multirow{2}{*}{$\mathtt{LLaMA-7B}$}   & \textbf{chip2}      & -8.66E-07 & 0.0155 & 0.083   & -0.082  \\
& \textbf{longform}   & -1.95E-06 & 0.0172 & 0.0859  & -0.085  \\
\midrule
\multirow{2}{*}{$\mathtt{LLaMA-13B}$}  & \textbf{chip2}      & 1.69E-06  & 0.0106 & 0.0762  & -0.0767 \\
& \textbf{longform}   & -1.47E-06 & 0.0173 & 0.0767  & -0.0747 \\
\midrule
\multirow{2}{*}{$\mathtt{LLaMA-33B}$}  & \textbf{chip2}      & -2.39E-07 & 0.0095 & 0.0688  & -0.0703 \\
 & \textbf{longform}   & -8.80E-08 & 0.0075 & 0.0703  & -0.0708 \\
\midrule
\multirow{2}{*}{$\mathtt{LLaMA-65B}$}  & \textbf{chip2}      & -5.07E-07 & 0.0083 & 0.062   & -0.063  \\
  & \textbf{longform}   & -7.91E-08 & 0.0097 & 0.0635  & -0.064  \\
\midrule
\multirow{2}{*}{$\mathtt{LLaMA2-70B}$} & \textbf{chip2}      & 3.63E-07  & 0.0053 & 0.0396  & -0.0393 \\
 & \textbf{longform}   & -1.16E-07 & 0.0053 & 0.0427  & -0.043  \\ 
\bottomrule
\end{tabular}
\end{table}

\subsection{Why Use Multiplication Factor of Standard Deviation}
\label{app:std}

This decision to use multiplication factors of std is based on a few observations: (1) the task vector parameters are typically normally distributed with almost zero means (see table below), implying that the pretrained parameters have not changed much on average and some specific parameters get huge updates while others just accumulate SGD noise. (2) This std of task vectors can differ a lot based on the model size and the dataset. (3) We only care about the top-k fraction of the parameters (say top-20\%) that lie outside the first standard deviation (> $\sigma$), i.e., that has a magnitude greater than $\sigma$. Hence, given this different scale of top-k task vector parameters across different model sizes, tasks, etc., the standard deviation serves as a nice unifying scale that provides us with a constant set of values to try for $\alpha$, making this process simpler.

In Table~\ref{tab:statistics}, we provide the mean, standard deviation, maximum, and minimum values of the task vectors for models of different sizes and datasets. We observe that std, max, and min values change as the model size changes. For example, for the 3B model, the std is ~0.13 while for the 70B model, the std is ~0.009. Hence, we use $\alpha$ * $\sigma$ as it allows us to try hyperparameters in the correct range. However, we agree that there might be other ways to go about selecting $\alpha$, for example, learning on a small dataset.

% \begin{table}[t!]
% \centering

% \captionsetup{type=table}
% \caption{\label{tab:timing} Wall clock time for \methodshort{} and Original checkpoints for two different experimental setups.}

% \begin{tabular}{lcccc}
% \toprule
% \textbf{Model ($\downarrow$)} & \multicolumn{2}{c}{\textbf{Internet $\rightarrow$ Local (seconds)}} & \multicolumn{2}{c}{\textbf{CPU $\rightarrow$ GPU (milliseconds)}} \\
% \cmidrule(lr){2-3} \cmidrule(lr){4-5} 
% & \textbf{Original} & \methodshortbold{} & \textbf{Original} & \methodshortbold{} \\
% \midrule

% $\mathtt{LLaMA-7B}$  & $11.21_{2.44}$ & $1.16_{0.04}$ & $134.28_{4.76}$ & $11.23_{5.22}$ \\
% $\mathtt{LLaMA-13B}$ & $16.85_{3.83}$ & $1.75_{0.3}$ & $186.6_{5.42}$ & $23.09_{0.78}$ \\
% $\mathtt{LLaMA-33B}$ & $32.31_{6.76}$ & $2.46_{0.12}$ & $307.29_{55.59}$ & $18.0_{4.34}$ \\
% $\mathtt{LLaMA-65B}$ & $83.17_{9.14}$ & $2.59_{0.14}$ & $475.26_{66.51}$ & $18.6_{5.67}$ \\ 
% \bottomrule
% \end{tabular}

% \end{table}

\section{Additional Results}

\subsection{Comparision With Other Additional Pruning Methods}

We performed additional experiments in a setting similar to Table-\ref{tab:llama2}, where we worked with the Llama-2~\citep{touvron2023llama} 70B model and learned qLora~\citep{dettmers2023qlora} modules of rank 64. We then compressed these parameter updates using ComPEFT, STC, BitDelta~\citep{liu2024bitdelta}, and DAREx~\citep{deng2024dare} methods. Note that the BitDelta method has two variants. The first variant does not perform any additional training for the scale parameter (referred to as “No Training”). In the “BitDelta (No Training)” setting, the scale parameter ($\alpha$) is set to the mean value of all the parameters in the task vector/delta weights. The “BitDelta (Training)” variant learns the scale parameter ($\alpha$) via SGD and hence is not directly comparable with our ComPEFT which requires no additional training. For the DAREx method we use the DAREx-q ($1/q_v$) variant, which uses labelled data to select the inverse scaling parameter ($q_v$) for each per-layer separately after pruning. We DAREx, we use sparsity levels of 95\% and 99\% as used in their paper. The results for the experiments are provided below along the average sizes of the compressed parameters across all the tasks.

\begin{table}[th!]
    \centering
    \vspace{10pt}
    \caption{\label{tab:pruningbaseline} Comparing ComPEFT with other additional Pruning methods.}
    \resizebox{\linewidth}{!}{  
    \begin{tabular}{lccccccc}
    \toprule
        \textbf{Dataset} & \textbf{Original} & \textbf{ComPEFT} & \textbf{STC} & 
        \multicolumn{2}{c}{\textbf{BitDelta}} & \multicolumn{2}{c}{\textbf{DAREx-$q_v$}} \\
        \cmidrule(lr){5-6} \cmidrule(lr){7-8}
        & & & & \textbf{No Training} & \textbf{Training} & \textbf{p=0.95} & \textbf{p=0.99} \\
        \midrule
        \textbf{alpaca-clean} & 67.13 & 67.56 & 66.57 & 66.27 & 67.43 & 65.85 & 39.57 \\
        \textbf{chip2} & 65.18 & 67 & 64.54 & 64.31 & 67.31 & 63.94 & 50.18 \\ 
        \textbf{longform} & 67.63 & 68.5 & 67.02 & 66.15 & 68.61 & 66.14 & 44.32 \\
        \textbf{oasst1} & 66.89 & 67.39 & 66.15 & 65.38 & 67.11 & 65.48 & 45.82 \\
        \textbf{self-instruct} & 62.36 & 67.18 & 61.94 & 61.52 & 66.82 & 61.97 & 49.39 \\
        \midrule
        \textbf{Average} & 65.84 & 67.53 & 65.24 & 64.73 & 67.46 & 64.68 & 45.86 \\ 
        \textbf{Size} & 1.58GB & 56MB & 56MB & 99MB & 99MB & 395MB & 79MB \\
        \bottomrule
    \end{tabular}
    }
\end{table}

From these results, we can clearly see that: (1) ComPEFT performs better than these baseline. (2) DAREx (p=0.95) and BitDelta(No Training) show slight performance loss compared to the original checkpoint while DAREx (p=0.99) results in a huge drop. This is in line with the results presented in their papers. (3) BitDelta (Training) performs similar to ComPEFT, however, this method learns the scalar ($\alpha$) which requires both forward and backward passes and hence more GPU memory. (4) Note that BitDelta (No Training) sets the scalar ($\alpha$) as the mean of all the values in the task vector. It is very similar to STC which also uses the mean value as the scalar. However, they have a critical difference which is that STC also performs sparsification before performing quantization. Hence, in BitDelta the values are (+a, -a) while in STC the values are like (+b,0,-b). We note that STC performs slightly better than BitDelta (No Training), we believe that this is due to the sparsification step which removes redundant parameters which add noise. Similar phenomenon is also observed in TIES-Merging~\citep{yadav2023ties-merging}. Lastly, we also report the storage size for the compressed checkpoints where we use different methods to store them. We use golomb coding for ComPEFT/STC, bitmask for Bitdelta, and coo\_sparse matrix for DAREx method. The results demonstrate that ComPEFT yields better performance/size trade-off compared to most of these other methods.

\subsection{Comparisons with Advanced PEFT Methods}

\begin{table}[th!]
    \centering
    \caption{\label{tab:peft} Comparison with other PEFT methods}
    \begin{tabular}{lcccc}
    \toprule
        \textbf{Dataset} & \textbf{LORA} & \textbf{ComLORA} & \textbf{DORA} & \textbf{ComDORA} \\
        \midrule
        \textbf{alpaca-clean} & 67.13 & 67.56 & 68.42 & 69.78 \\
        \textbf{chip2} & 65.18 & 67 & 67.21 & 68.32 \\ 
        \textbf{longform} & 67.63 & 68.5 & 69.36 & 68.92 \\
        \textbf{oasst1} & 66.89 & 67.39 & 68.89 & 67.63 \\ 
        \textbf{self-instruct} & 62.36 & 67.18 & 65.26 & 66.31 \\
        \midrule
        \textbf{Average} & 65.84 & 67.53 & 67.83 & 68.19 \\
        \textbf{Size} & 1.58GB & 56MB & 1.59GB & 57MB \\
        \bottomrule
    \end{tabular}
\end{table}

We conducted some additional experiments with some other PEFt methods like DoRA~\citep{liu2024dora}. For the experimental setting in Table-\ref{tab:llama2} with rank 64 LoRA on the Llama-2 70B model. We performed additional experiments with DoRA of rank 64 and then compressed them using ComPEFT and reported the results. We omitted VeRA~\citep{kopiczko2023vera} methods as based on the DoRA paper VeRA typically performs worse than both LoRA and DoRA. Lastly, we omitted HiRA~\citep{huang2025hira} as the method as due to its recency its code is not available. In Table~\ref{tab:peft} we present our results. Similar to our other finding, we observe that ComPEFT can also compress DoRA checkpoints to a great extent while preserving performance. Moreover, ComDoRA checkpoints slightly outperform ComLoRA’s performance.

\subsection{Comparison of Compressed Lora With Lower Rank Lora Modules}

We perform additional experiments to compare the compressed LoRA modules with lower rank lora module which inherently have smaller sizes as compression can be achieved on smaller rank. We opt for the experimental setting from Table-\ref{tab:llama2} where we work with the Llama-2 70B model. We perform experiments with rank 32 and 8 the results of which are attached below along with storage sizes.

From the results in Table~\ref{tab:rankreduction}, we can see that: (1) at rank 32 there is a slight drop in performance compared to rank 64. We can compress the rank checkpoint as well by >25x (2) At rank 8, we see a significant drop in performance from 65.84 to 63.77. Moreover, ComPEFT can compress rank 8 lora as well by >25x. (3) for both rank 32 and 8, ComLoRA performs better than the original checkpoints. These results help us to conclude that the observed benefits in compression and performance improvements stem from ComPEFT as opposed to the overparameterized LoRA adapter.

\begin{table}[th!]
    \centering
    \vspace{10pt}
    \caption{\label{tab:rankreduction} Comparing ComPEFT with smaller rank LoRA modules.}
    \resizebox{\linewidth}{!}{  
    \begin{tabular}{lcccccc}
    \toprule
        \textbf{Dataset} & \textbf{Lora(r=64)} & \textbf{ComLora(r=64)} & \textbf{Lora(r=32)} & \textbf{ComLora(r=32)} & \textbf{Lora(r=8)} & \textbf{ComLora(r=8)} \\ 
        \midrule
        \textbf{alpaca-clean} & 67.13 & 67.56 & 66.98 & 67.24 & 64.82 & 65.27 \\ 
        \textbf{chip2} & 65.18 & 67 & 65.24 & 66.75 & 63.35 & 65.18 \\ 
        \textbf{longform} & 67.63 & 68.5 & 67.14 & 68.12 & 65.16 & 66.74 \\ 
        \textbf{oasst1} & 66.89 & 67.39 & 65.42 & 66.92 & 64.21 & 65.56 \\ 
        \textbf{self-instruct} & 62.36 & 67.18 & 62.68 & 67.48 & 61.32 & 65.81 \\ 
        \midrule
        \textbf{Average} & 65.84 & 67.53 & 65.49 & 67.30 & 63.77 & 65.71 \\ 
        \textbf{Size} & 1.58GB & 56MB & 790MB & 28MB & 197MB & 7MB \\ 
        \bottomrule
    \end{tabular}
    }
\end{table}

\begin{table*}[t!]
\centering
\caption{\label{tab:llama_val} We present the performance$_{\text{(Storage Size in GB)}}$ on MMLU Validation for the compressed QLoRA models.}

\begin{tabular}{lcccccccc}
\toprule

\textbf{$\mathtt{Dataset}$ ($\downarrow$)} & \multicolumn{4}{c}{\methodshort{}} \\
\cmidrule(lr){2-5}

& $\mathtt{7B}$ & $\mathtt{13B}$ & $\mathtt{33B}$ & $\mathtt{65B}$ \\
\midrule

\textbf{Self-Instruct} & 35.62 & 47.52 & 55.11 & 62.13  \\
\textbf{Longform} & 31.89 & 47.80 & 55.31 & 62.27  \\
\textbf{Chip2} & 33.49 & 47.15 & 55.02 & 62.21  \\
\textbf{HH-RLHF} & 32.37 & 47.19 & 54.78 & 62.06  \\
\textbf{Unnatural Instruct} & 42.41 & 49.62 & 56.28 & 62.15  \\
\textbf{Guanaco} & 33.92 & 49.52 & 55.35 & 62.00  \\
\textbf{Alpaca} & 39.82 & 49.00 & 55.91 & 62.37  \\
\textbf{FLAN v2} & 43.93 & 50.86 & 56.97 & 63.77  \\
\midrule
\textbf{Average} & 37.88 & 48.58 & 55.59 & 62.37  \\

\bottomrule
\end{tabular}
\end{table*}

\begin{table}[t!]
\centering
\captionsetup{type=table}
\caption{\label{tab:goodzs_peft_val} Validation set performance$_\text{(Storage Size in MB)}$ averaged over seven GLUE tasks when compressing (IA)$^3$ and LoRA modules on different base models.}

\begin{tabular}{lccccccccc}
\toprule

\textbf{$\mathtt{Method}$ ($\downarrow$)} & \multicolumn{2}{c}{$\mathtt{T5-Base}$}  & \multicolumn{2}{c}{$\mathtt{T5-Large}$} & \multicolumn{2}{c}{$\mathtt{T0-3B}$} \\
 \cmidrule(lr){2-3} \cmidrule(lr){4-5} \cmidrule(lr){6-7} \cmidrule(lr){8-9}

 & $\mathtt{(IA)^3}$ & $\mathtt{LoRA}$ & $\mathtt{(IA)^3}$ & $\mathtt{LoRA}$ & $\mathtt{(IA)^3}$ & $\mathtt{LoRA}$ \\
 \midrule

\textsc{Original} & 81.25 &	81.94 & 85.08 &	86.21 & 87.71 & 89.94 \\
\textsc{\methodshort{}} & 81.04 & 80.96 & 85.28 & 86.54 & 89.14 & 89.95 \\
\midrule
\textsc{Improvement} & -0.21 & -0.98 & 0.2 & 0.33 & 1.43 & 0.01 \\

\bottomrule
\end{tabular}
\end{table}

\subsection{Validation Set Results}
\label{sec:app_validation}
In Table~\ref{tab:llama_val} and~\ref{tab:goodzs_peft_val}, we provide the validation set results for our main compression experiments on $\mathtt{LLaMA, T5, T0}$ experiments from Section~\ref{sec:peft_llama} and~\ref{sec:peft_small} respectively.

\subsection{Full Results for Compositional Generalization}
\label{sec:app_lorahub}
In Table~\ref{tab:lorahub_full}, we present the Zeroshot, ICL, LoraHub, and \methodshort{} results for each of the BBH tasks.

\subsection{Individual Task Results}
\label{sec:app_ind_results}
We present the task level validation and test set results along with model sizes of (IA)$^3$, LoRA, and full finetuning for T5-base (Table~\ref{tab:t5-base}), T5-large (Table~\ref{tab:t5-large}), T0-3B (Table~\ref{tab:bigscience/T0_3B}).

\subsection{Compressing Model With Smaller Models with Bad ZeroShot Performance}
\label{sec:app_bad_zs}
We present the task level validation and test set results along with model sizes for (IA)$^3$, LoRA, and full finetuning for BERT-base (Table~\ref{tab:bert-base-uncased}), BERT-large (Table~\ref{tab:bert-large-uncased}), RoBERTa-base (Table~\ref{tab:roberta-base}), RoBERTa-large (Table~\ref{tab:roberta-large}), T5-v1.1-base (Table~\ref{tab:google/t5-v1_1-base}), and T5-v1.1-large (Table~\ref{tab:google/t5-v1_1-large}). These models are only trained using the pretraining objective and are not multitask-trained. Hence, these models have very bad zero/few-shot performance and always require explicit finetuning to perform well on any downstream tasks. We observe that for the LoRA method, the performance of \methodshort{} is similar to the uncompressed full models while being smaller in size. This hints at the fact that the intrinsic dimensionality of the LoRA adaptation is much smaller compared to the number of parameters in the LoRA module. However, for (IA)$^3$ method, the performance drop is more, we believe that two reasons for this are: (1) The models are not good zero/few-shot models, and (2) (IA)$^3$ adds very few parameters to perform a multiplicative operation on the activations. Therefore, the loss landscape is not as smooth as for good zershot models, and due to this IA3 has to scale different activations in a very different manner to learn the task. Hence, compressing (IA)$^3$ to sparse sign-vector and a constant is not feasible. Whereas, In the case of Lora the updates are added and hence their impact on the final value of the parameter is not huge as the maximum of the LoRA parameter is still very small compared to the base model's parameter value.

\begin{table*}
\caption{\textbf{Task level results:} Average performance over 5 seed for LoraHub and \methodshort{} for compositional generalization on Big-Bench-Hard.}
\label{tab:lorahub_full}
\resizebox{\linewidth}{!}{  
\begin{tabular}{lcccccc}
\toprule
\textbf{Task} & \textbf{Zeroshot} & \textbf{ICL} & \textbf{LoraHub (Avg)} & \textbf{\methodshort{} (Avg)} & \textbf{LoraHub (Best)} & \textbf{\methodshort{} (Best)} \\

\midrule
\textbf{Logical Deduction Three Objects} & 0.0 & 51.3 & 41.9 & 28.4 & 51.3 & 48.0 \\
\textbf{Tracking Shuffled Objects Five Objects} & 12.0 & 12.0 & 9.6 & 11.3 & 12.0 & 12.0 \\
\textbf{Web Of Lies} & 54.0 & 54.0 & 28.1 & 41.7 & 49.3 & 56.0 \\
\textbf{Tracking Shuffled Objects Seven Objects} & 6.7 & 6.7 & 5.3 & 6.7 & 6.7 & 6.7 \\
\textbf{Date Understanding} & 15.3 & 22.7 & 39.5 & 29.1 & 42.0 & 38.7 \\
\textbf{Navigate} & 47.3 & 44.0 & 48.4 & 38.5 & 50.7 & 50.0 \\
\textbf{Multistep Arithmetic Two} & 0.7 & 0.7 & 0.7 & 0.5 & 1.3 & 0.7 \\
\textbf{Boolean Expressions} & 54.0 & 58.7 & 55.9 & 55.7 & 57.3 & 61.3 \\
\textbf{Hyperbaton} & 6.7 & 74.0 & 55.2 & 49.9 & 65.3 & 67.3 \\
\textbf{Tracking Shuffled Objects Three Objects} & 24.7 & 30.7 & 26.7 & 21.6 & 29.3 & 24.7 \\
\textbf{Sports Understanding} & 56.0 & 56.0 & 46.4 & 53.1 & 54.7 & 58.0 \\
\textbf{Logical Deduction Seven Objects} & 12.7 & 42.0 & 35.5 & 37.6 & 40.0 & 40.0 \\
\textbf{Causal Judgement} & 57.5 & 56.3 & 40.7 & 49.2 & 58.6 & 57.5 \\
\textbf{Penguins In A Table} & 43.5 & 39.1 & 36.1 & 44.3 & 45.7 & 47.8 \\
\textbf{Geometric Shapes} & 6.7 & 18.7 & 9.6 & 7.3 & 19.3 & 9.3 \\
\textbf{Reasoning About Colored Objects} & 32.0 & 38.7 & 38.0 & 40.8 & 39.3 & 44.0 \\
\textbf{Dyck Languages} & 1.3 & 2.7 & 1.1 & 0.7 & 1.3 & 1.3 \\
\textbf{Disambiguation Qa} & 0.0 & 69.3 & 14.3 & 6.5 & 51.3 & 29.3 \\
\textbf{Salient Translation Error Detection} & 37.3 & 46.0 & 31.3 & 38.5 & 44.7 & 43.3 \\
\textbf{Movie Recommendation} & 62.7 & 52.7 & 61.1 & 58.0 & 67.3 & 62.0 \\
\textbf{Snarks} & 50.0 & 55.1 & 49.2 & 50.0 & 50.0 & 50.0 \\
\textbf{Formal Fallacies} & 51.3 & 58.0 & 41.3 & 41.1 & 52.7 & 51.3 \\
\textbf{Logical Deduction Five Objects} & 21.3 & 40.0 & 33.6 & 36.3 & 36.7 & 42.0 \\
\textbf{Temporal Sequences} & 16.7 & 26.7 & 18.7 & 19.5 & 20.0 & 21.3 \\
\textbf{Word Sorting} & 1.3 & 0.7 & 1.2 & 1.3 & 1.3 & 1.3 \\
\textbf{Ruin Names} & 23.3 & 18.7 & 18.0 & 22.4 & 23.3 & 23.3 \\
\textbf{Object Counting} & 34.7 & 32.0 & 35.5 & 35.3 & 36.7 & 36.0 \\
\midrule
\textbf{Average} & 27.0 & 37.3 & 30.5 & 30.6 & 37.3 & 36.4 \\
\bottomrule
\end{tabular}
}
\end{table*}

\begin{table*}
\centering
\centering 
\caption{Validation and Test set performance along with storage size in MB for bert-base-uncased Model, for (IA)$^3$, LoRA and Full model finetuning.}
\label{tab:bert-base-uncased}
\begin{tabular}{llcccc}
\toprule
 &  & \textbf{\textbf{Original (Val)}} & \textbf{\textbf{Original (Test)}} & \textbf{\methodshort{} (Val)} & \textbf{\methodshort{} (Test)} \\
\textbf{PEFT} & \textbf{Task} &  &  &  &  \\
\midrule
\multirow[t]{7}{*}{\textbf{full}} & \textbf{mnli} & $84.7$ & $83.1_{(208.8)}$ & $84.3$ & $82.6_{(12.0)}$ \\
\textbf{} & \textbf{mrpc} & $86.8$ & $97.5_{(208.8)}$ & $86.8$ & $97.0_{(19.6)}$ \\
\textbf{} & \textbf{qnli} & $91.9$ & $91.0_{(208.8)}$ & $91.9$ & $90.9_{(12.0)}$ \\
\textbf{} & \textbf{qqp} & $90.1$ & $90.3_{(208.8)}$ & $90.0$ & $90.0_{(15.4)}$ \\
\textbf{} & \textbf{rte} & $66.4$ & $94.5_{(208.8)}$ & $69.0$ & $94.0_{(7.4)}$ \\
\textbf{} & \textbf{sst2} & $91.1$ & $97.0_{(208.8)}$ & $92.2$ & $96.0_{(7.4)}$ \\
\textbf{} & \textbf{wnli} & $56.3$ & $57.0_{(208.8)}$ & $56.3$ & $57.0_{(4.4)}$ \\
\midrule
\multirow[t]{7}{*}{\textbf{ia3}} & \textbf{mnli} & $79.2$ & $78.9_{(0.1)}$ & $57.6$ & $56.2_{(0.0)}$ \\
\textbf{} & \textbf{mrpc} & $84.6$ & $94.5_{(0.1)}$ & $31.6$ & $34.5_{(0.0)}$ \\
\textbf{} & \textbf{qnli} & $87.9$ & $87.5_{(0.1)}$ & $49.3$ & $49.6_{(0.0)}$ \\
\textbf{} & \textbf{qqp} & $84.6$ & $84.4_{(0.1)}$ & $63.2$ & $63.2_{(0.0)}$ \\
\textbf{} & \textbf{rte} & $59.2$ & $73.5_{(0.1)}$ & $52.7$ & $55.5_{(0.0)}$ \\
\textbf{} & \textbf{sst2} & $91.5$ & $91.0_{(0.1)}$ & $49.1$ & $45.5_{(0.0)}$ \\
\textbf{} & \textbf{wnli} & $54.9$ & $57.0_{(0.1)}$ & $56.3$ & $57.0_{(0.0)}$ \\
\midrule
\multirow[t]{7}{*}{\textbf{lora}} & \textbf{mnli} & $82.5$ & $81.4_{(2.6)}$ & $76.6$ & $76.9_{(0.2)}$ \\
\textbf{} & \textbf{mrpc} & $86.3$ & $97.5_{(2.6)}$ & $82.8$ & $94.0_{(0.2)}$ \\
\textbf{} & \textbf{qnli} & $91.7$ & $91.0_{(2.6)}$ & $90.8$ & $90.1_{(0.2)}$ \\
\textbf{} & \textbf{qqp} & $89.4$ & $89.5_{(2.6)}$ & $87.4$ & $87.5_{(0.2)}$ \\
\textbf{} & \textbf{rte} & $61.7$ & $68.5_{(2.6)}$ & $58.8$ & $66.0_{(0.2)}$ \\
\textbf{} & \textbf{sst2} & $92.4$ & $92.5_{(2.6)}$ & $91.6$ & $91.5_{(0.2)}$ \\
\textbf{} & \textbf{wnli} & $56.3$ & $57.0_{(2.6)}$ & $59.2$ & $56.0_{(0.1)}$ \\
\bottomrule
\end{tabular}
\end{table*}

\begin{table*}
\centering 
\caption{Validation and Test set performance along with storage size in MB for bert-large-uncased Model, for (IA)$^3$, LoRA and Full model finetuning.}
\label{tab:bert-large-uncased}
\begin{tabular}{llcccc}
\toprule
 &  & \textbf{Original (Val)} & \textbf{Original (Test)} & \textbf{ \methodshort{} (Val)} & \textbf{ \methodshort{} (Test)} \\
\textbf{PEFT} & \textbf{Task} &  &  &  &  \\
\midrule
\multirow[t]{7}{*}{\textbf{full}} & \textbf{mnli} & $85.4$ & $84.0_{(639.2)}$ & $85.5$ & $83.7_{(59.9)}$ \\
\textbf{} & \textbf{mrpc} & $88.2$ & $97.5_{(639.2)}$ & $88.5$ & $97.5_{(47.2)}$ \\
\textbf{} & \textbf{qnli} & $91.0$ & $89.4_{(639.2)}$ & $91.2$ & $89.6_{(36.8)}$ \\
\textbf{} & \textbf{qqp} & $88.6$ & $88.5_{(639.2)}$ & $88.4$ & $88.6_{(36.8)}$ \\
\textbf{} & \textbf{rte} & $71.5$ & $94.0_{(639.2)}$ & $70.4$ & $92.5_{(22.7)}$ \\
\textbf{} & \textbf{sst2} & $92.9$ & $93.5_{(639.2)}$ & $93.5$ & $92.5_{(36.8)}$ \\
\textbf{} & \textbf{wnli} & $56.3$ & $57.0_{(639.2)}$ & $57.8$ & $58.0_{(13.4)}$ \\
\midrule
\multirow[t]{7}{*}{\textbf{ia3}} & \textbf{mnli} & $82.4$ & $81.9_{(0.3)}$ & $59.5$ & $59.6_{(0.0)}$ \\
\textbf{} & \textbf{mrpc} & $84.6$ & $96.0_{(0.3)}$ & $31.6$ & $34.5_{(0.0)}$ \\
\textbf{} & \textbf{qnli} & $88.6$ & $87.6_{(0.3)}$ & $59.6$ & $59.9_{(0.0)}$ \\
\textbf{} & \textbf{qqp} & $87.7$ & $87.4_{(0.3)}$ & $73.2$ & $73.8_{(0.0)}$ \\
\textbf{} & \textbf{rte} & $58.8$ & $72.5_{(0.3)}$ & $52.7$ & $55.5_{(0.0)}$ \\
\textbf{} & \textbf{sst2} & $92.3$ & $88.0_{(0.3)}$ & $49.1$ & $45.5_{(0.0)}$ \\
\textbf{} & \textbf{wnli} & $60.6$ & $55.0_{(0.3)}$ & $56.3$ & $57.0_{(0.0)}$ \\
\midrule
\multirow[t]{7}{*}{\textbf{lora}} & \textbf{mnli} & $83.6$ & $82.9_{(6.8)}$ & $76.4$ & $74.8_{(0.6)}$ \\
\textbf{} & \textbf{mrpc} & $88.7$ & $94.0_{(6.8)}$ & $87.0$ & $92.0_{(0.5)}$ \\
\textbf{} & \textbf{qnli} & $88.6$ & $86.9_{(6.8)}$ & $82.6$ & $81.1_{(0.6)}$ \\
\textbf{} & \textbf{qqp} & $87.0$ & $87.1_{(6.8)}$ & $76.3$ & $76.9_{(0.6)}$ \\
\textbf{} & \textbf{rte} & $59.9$ & $77.0_{(6.8)}$ & $59.6$ & $71.0_{(0.6)}$ \\
\textbf{} & \textbf{sst2} & $93.7$ & $94.0_{(6.8)}$ & $93.6$ & $93.0_{(0.4)}$ \\
\textbf{} & \textbf{wnli} & $56.3$ & $57.0_{(6.8)}$ & $57.8$ & $56.0_{(0.2)}$ \\
\bottomrule
\end{tabular}
\end{table*}

\begin{table*}
\centering 
\caption{Validation and Test set performance along with storage size in MB for roberta-base Model, for (IA)$^3$, LoRA and Full model finetuning.}
\label{tab:roberta-base}
\begin{tabular}{llcccc}
\toprule
 &  & \textbf{Original (Val)} & \textbf{Original (Test)} & \textbf{ \methodshort{} (Val)} & \textbf{ \methodshort{} (Test)} \\
\textbf{PEFT} & \textbf{Task} &  &  &  &  \\
\midrule
\multirow[t]{7}{*}{\textbf{full}} & \textbf{mnli} & $86.4$ & $86.4_{(237.8)}$ & $86.6$ & $86.2_{(13.7)}$ \\
\textbf{} & \textbf{mrpc} & $87.0$ & $89.0_{(237.8)}$ & $86.0$ & $84.5_{(8.4)}$ \\
\textbf{} & \textbf{qnli} & $91.8$ & $91.2_{(237.8)}$ & $91.7$ & $91.0_{(17.6)}$ \\
\textbf{} & \textbf{qqp} & $89.1$ & $89.4_{(237.8)}$ & $89.1$ & $89.2_{(17.6)}$ \\
\textbf{} & \textbf{rte} & $75.4$ & $91.5_{(237.8)}$ & $78.0$ & $93.0_{(22.3)}$ \\
\textbf{} & \textbf{sst2} & $95.2$ & $95.2_{(237.8)}$ & $94.2$ & $94.0_{(8.4)}$ \\
\textbf{} & \textbf{wnli} & $56.3$ & $56.0_{(237.8)}$ & $56.3$ & $45.0_{(5.0)}$ \\
\midrule
\multirow[t]{7}{*}{\textbf{ia3}} & \textbf{mnli} & $84.1$ & $83.4_{(1.2)}$ & $43.0$ & $44.7_{(0.1)}$ \\
\textbf{} & \textbf{mrpc} & $88.7$ & $98.0_{(1.2)}$ & $71.6$ & $70.0_{(0.1)}$ \\
\textbf{} & \textbf{qnli} & $89.7$ & $88.9_{(1.2)}$ & $50.7$ & $50.4_{(0.1)}$ \\
\textbf{} & \textbf{qqp} & $87.0$ & $87.1_{(1.2)}$ & $80.9$ & $80.8_{(0.1)}$ \\
\textbf{} & \textbf{rte} & $73.3$ & $93.0_{(1.2)}$ & $54.9$ & $54.5_{(0.1)}$ \\
\textbf{} & \textbf{sst2} & $93.5$ & $92.0_{(1.2)}$ & $76.3$ & $72.5_{(0.1)}$ \\
\textbf{} & \textbf{wnli} & $56.3$ & $57.0_{(1.2)}$ & $56.3$ & $57.0_{(0.0)}$ \\
\midrule
\multirow[t]{7}{*}{\textbf{lora}} & \textbf{mnli} & $87.0$ & $86.1_{(3.7)}$ & $86.2$ & $85.6_{(0.3)}$ \\
\textbf{} & \textbf{mrpc} & $89.5$ & $98.5_{(3.7)}$ & $88.5$ & $97.0_{(0.3)}$ \\
\textbf{} & \textbf{qnli} & $91.1$ & $92.3_{(3.7)}$ & $90.0$ & $89.9_{(0.3)}$ \\
\textbf{} & \textbf{qqp} & $88.8$ & $88.8_{(3.7)}$ & $88.2$ & $88.4_{(0.3)}$ \\
\textbf{} & \textbf{rte} & $79.4$ & $97.0_{(3.7)}$ & $79.4$ & $96.0_{(0.3)}$ \\
\textbf{} & \textbf{sst2} & $94.2$ & $95.0_{(3.7)}$ & $93.1$ & $94.0_{(0.3)}$ \\
\textbf{} & \textbf{wnli} & $56.3$ & $57.0_{(3.7)}$ & $56.3$ & $57.0_{(0.1)}$ \\

\bottomrule
\end{tabular}
\end{table*}

\begin{table*}
\centering 
\caption{Validation and Test set performance along with storage size in MB for roberta-large Model, for (IA)$^3$, LoRA and Full model finetuning.}
\label{tab:roberta-large}
\begin{tabular}{llcccc}
\toprule
 &  & \textbf{Original (Val)} & \textbf{Original (Test)} & \textbf{ \methodshort{} (Val)} & \textbf{ \methodshort{} (Test)} \\
\textbf{PEFT} & \textbf{Task} &  &  &  &  \\
\midrule
\multirow[t]{7}{*}{\textbf{full}} & \textbf{mnli} & $90.6$ & $89.3_{(677.8)}$ & $90.4$ & $89.4_{(50.0)}$ \\
\textbf{} & \textbf{mrpc} & $89.2$ & $97.5_{(677.8)}$ & $89.2$ & $97.5_{(24.1)}$ \\
\textbf{} & \textbf{qnli} & $93.6$ & $92.9_{(677.8)}$ & $93.5$ & $93.3_{(63.5)}$ \\
\textbf{} & \textbf{qqp} & $90.1$ & $89.9_{(677.8)}$ & $90.1$ & $89.9_{(50.0)}$ \\
\textbf{} & \textbf{rte} & $85.6$ & $98.5_{(677.8)}$ & $85.2$ & $98.0_{(50.0)}$ \\
\textbf{} & \textbf{sst2} & $95.4$ & $95.2_{(677.8)}$ & $96.3$ & $95.0_{(50.0)}$ \\
\textbf{} & \textbf{wnli} & $56.3$ & $57.0_{(677.8)}$ & $57.8$ & $61.0_{(63.5)}$ \\
\midrule
\multirow[t]{7}{*}{\textbf{ia3}} & \textbf{mnli} & $89.5$ & $88.5_{(2.3)}$ & $36.5$ & $35.5_{(0.2)}$ \\
\textbf{} & \textbf{mrpc} & $86.8$ & $86.5_{(2.3)}$ & $68.4$ & $65.5_{(0.0)}$ \\
\textbf{} & \textbf{qnli} & $92.3$ & $92.1_{(2.3)}$ & $51.6$ & $50.4_{(0.1)}$ \\
\textbf{} & \textbf{qqp} & $88.5$ & $87.9_{(2.3)}$ & $63.2$ & $63.2_{(0.0)}$ \\
\textbf{} & \textbf{rte} & $80.1$ & $94.0_{(2.3)}$ & $54.2$ & $57.0_{(0.0)}$ \\
\textbf{} & \textbf{sst2} & $94.3$ & $93.0_{(2.3)}$ & $55.2$ & $59.5_{(0.2)}$ \\
\textbf{} & \textbf{wnli} & $56.3$ & $57.0_{(2.3)}$ & $60.6$ & $51.0_{(0.1)}$ \\
\midrule
\multirow[t]{7}{*}{\textbf{lora}} & \textbf{mnli} & $89.9$ & $89.3_{(8.8)}$ & $85.0$ & $83.8_{(0.6)}$ \\
\textbf{} & \textbf{mrpc} & $90.0$ & $93.5_{(8.8)}$ & $90.4$ & $90.0_{(0.8)}$ \\
\textbf{} & \textbf{qnli} & $93.4$ & $92.9_{(8.8)}$ & $91.0$ & $90.3_{(0.8)}$ \\
\textbf{} & \textbf{qqp} & $89.1$ & $88.9_{(8.8)}$ & $86.5$ & $85.9_{(0.8)}$ \\
\textbf{} & \textbf{rte} & $80.5$ & $93.5_{(8.8)}$ & $79.1$ & $89.5_{(0.8)}$ \\
\textbf{} & \textbf{sst2} & $95.2$ & $93.0_{(8.8)}$ & $94.8$ & $90.5_{(0.8)}$ \\
\textbf{} & \textbf{wnli} & $56.3$ & $57.0_{(8.8)}$ & $56.3$ & $57.0_{(0.2)}$ \\

\bottomrule
\end{tabular}
\end{table*}

\begin{table*}
\centering 
\caption{Validation and Test set performance along with storage size in MB for t5-v1.1-base Model, for (IA)$^3$, LoRA and Full model finetuning.}
\label{tab:google/t5-v1_1-base}
\begin{tabular}{llcccc}
\toprule
 &  & \textbf{Original (Val)} & \textbf{Original (Test)} & \textbf{ \methodshort{} (Val)} & \textbf{ \methodshort{} (Test)} \\
\textbf{PEFT} & \textbf{Task} &  &  &  &  \\
\midrule
\multirow[t]{7}{*}{\textbf{full}} & \textbf{mnli} & $89.8$ & $89.8_{(472.2)}$ & $88.8$ & $89.0_{(34.9)}$ \\
\textbf{} & \textbf{mrpc} & $80.9$ & $74.5_{(472.2)}$ & $82.6$ & $75.5_{(34.9)}$ \\
\textbf{} & \textbf{qnli} & $88.0$ & $88.6_{(472.2)}$ & $86.7$ & $87.2_{(44.3)}$ \\
\textbf{} & \textbf{qqp} & $78.6$ & $78.9_{(472.2)}$ & $77.0$ & $77.6_{(27.2)}$ \\
\textbf{} & \textbf{rte} & $59.2$ & $49.0_{(472.2)}$ & $59.2$ & $61.0_{(9.9)}$ \\
\textbf{} & \textbf{sst2} & $93.4$ & $91.0_{(472.2)}$ & $93.8$ & $91.5_{(27.2)}$ \\
\textbf{} & \textbf{wnli} & $56.3$ & $47.0_{(472.2)}$ & $57.8$ & $49.0_{(44.3)}$ \\
\midrule
\multirow[t]{7}{*}{\textbf{ia3}} & \textbf{mnli} & $84.6$ & $83.8_{(0.2)}$ & $54.3$ & $54.4_{(0.0)}$ \\
\textbf{} & \textbf{mrpc} & $82.8$ & $81.5_{(0.2)}$ & $82.8$ & $78.5_{(0.0)}$ \\
\textbf{} & \textbf{qnli} & $85.2$ & $86.3_{(0.2)}$ & $60.7$ & $61.8_{(0.0)}$ \\
\textbf{} & \textbf{qqp} & $85.0$ & $85.4_{(0.2)}$ & $78.6$ & $78.8_{(0.0)}$ \\
\textbf{} & \textbf{rte} & $54.9$ & $49.0_{(0.2)}$ & $63.2$ & $62.5_{(0.0)}$ \\
\textbf{} & \textbf{sst2} & $92.3$ & $91.0_{(0.2)}$ & $89.0$ & $87.0_{(0.0)}$ \\
\textbf{} & \textbf{wnli} & $52.1$ & $57.0_{(0.2)}$ & $52.1$ & $57.0_{(0.0)}$ \\
\midrule
\multirow[t]{7}{*}{\textbf{lora}} & \textbf{mnli} & $66.9$ & $66.3_{(4.4)}$ & $56.6$ & $57.1_{(0.2)}$ \\
\textbf{} & \textbf{mrpc} & $72.1$ & $67.0_{(4.4)}$ & $68.4$ & $64.0_{(0.3)}$ \\
\textbf{} & \textbf{qnli} & $87.1$ & $88.8_{(4.4)}$ & $86.7$ & $88.5_{(0.3)}$ \\
\textbf{} & \textbf{qqp} & $78.3$ & $78.9_{(4.4)}$ & $72.0$ & $72.2_{(0.2)}$ \\
\textbf{} & \textbf{rte} & $55.2$ & $50.5_{(4.4)}$ & $53.1$ & $49.5_{(0.2)}$ \\
\textbf{} & \textbf{sst2} & $93.0$ & $91.5_{(4.4)}$ & $92.9$ & $90.5_{(0.3)}$ \\
\textbf{} & \textbf{wnli} & $56.3$ & $47.0_{(4.4)}$ & $78.9$ & $76.0_{(0.1)}$ \\

\bottomrule
\end{tabular}
\end{table*}

\begin{table*}
\centering 
\caption{Validation and Test set performance along with storage size in MB for t5-v1.1-large Model, for (IA)$^3$, LoRA and Full model finetuning.}
\label{tab:google/t5-v1_1-large}
\begin{tabular}{llcccc}
\toprule
 &  & \textbf{Original (Val)} & \textbf{Original (Test)} & \textbf{ \methodshort{} (Val)} & \textbf{ \methodshort{} (Test)} \\
\textbf{PEFT} & \textbf{Task} &  &  &  &  \\
\midrule
\multirow[t]{6}{*}{\textbf{full}} & \textbf{mrpc} & $86.3$ & $84.0_{(1493.7)}$ & $86.3$ & $85.0_{(110.3)}$ \\
\textbf{} & \textbf{qnli} & $94.0$ & $94.0_{(1493.7)}$ & $94.4$ & $94.9_{(140.0)}$ \\
\textbf{} & \textbf{qqp} & $90.2$ & $90.5_{(1493.7)}$ & $89.4$ & $89.5_{(140.0)}$ \\
\textbf{} & \textbf{rte} & $74.0$ & $76.0_{(1493.7)}$ & $75.4$ & $74.5_{(140.0)}$ \\
\textbf{} & \textbf{sst2} & $95.6$ & $93.0_{(1493.7)}$ & $95.4$ & $92.5_{(86.1)}$ \\
\textbf{} & \textbf{wnli} & $52.1$ & $57.0_{(1493.7)}$ & $52.1$ & $57.0_{(31.4)}$ \\
\midrule
\multirow[t]{7}{*}{\textbf{ia3}} & \textbf{mnli} & $92.0$ & $92.4_{(0.5)}$ & $54.3$ & $54.4_{(0.0)}$ \\
\textbf{} & \textbf{mrpc} & $90.9$ & $86.0_{(0.5)}$ & $77.9$ & $77.5_{(0.0)}$ \\
\textbf{} & \textbf{qnli} & $92.0$ & $92.3_{(0.5)}$ & $79.4$ & $78.1_{(0.0)}$ \\
\textbf{} & \textbf{qqp} & $87.2$ & $87.5_{(0.5)}$ & $78.7$ & $78.8_{(0.0)}$ \\
\textbf{} & \textbf{rte} & $69.7$ & $67.0_{(0.5)}$ & $69.0$ & $73.5_{(0.0)}$ \\
\textbf{} & \textbf{sst2} & $95.2$ & $93.0_{(0.5)}$ & $79.5$ & $81.0_{(0.0)}$ \\
\textbf{} & \textbf{wnli} & $52.1$ & $57.0_{(0.5)}$ & $52.1$ & $57.0_{(0.0)}$ \\
\midrule
\multirow[t]{7}{*}{\textbf{lora}} & \textbf{mnli} & $92.3$ & $93.2_{(11.8)}$ & $91.8$ & $92.4_{(0.9)}$ \\
\textbf{} & \textbf{mrpc} & $78.2$ & $74.0_{(11.8)}$ & $77.7$ & $76.0_{(0.9)}$ \\
\textbf{} & \textbf{qnli} & $90.4$ & $91.9_{(11.8)}$ & $87.0$ & $87.4_{(0.9)}$ \\
\textbf{} & \textbf{qqp} & $87.1$ & $87.6_{(11.8)}$ & $86.0$ & $86.8_{(0.9)}$ \\
\textbf{} & \textbf{rte} & $52.7$ & $56.0_{(11.8)}$ & $53.8$ & $45.5_{(0.4)}$ \\
\textbf{} & \textbf{sst2} & $93.9$ & $89.0_{(11.8)}$ & $61.4$ & $56.0_{(0.4)}$ \\
\textbf{} & \textbf{wnli} & $56.3$ & $47.0_{(11.8)}$ & $56.3$ & $47.0_{(0.4)}$ \\

\bottomrule
\end{tabular}
\end{table*}

\begin{table*}
\centering 
\caption{Validation and Test set performance along with storage size in MB for t5-base Model, for (IA)$^3$, LoRA and Full model finetuning.}
\label{tab:t5-base}
\begin{tabular}{llcccc}
\toprule
 &  & \textbf{Original (Val)} & \textbf{Original (Test)} & \textbf{ \methodshort{} (Val)} & \textbf{ \methodshort{} (Test)} \\
\textbf{PEFT} & \textbf{Task} &  &  &  &  \\
\midrule
\multirow[t]{7}{*}{\textbf{full}} & \textbf{mnli} & $91.2$ & $91.2_{(425.2)}$ & $89.9$ & $90.4_{(31.4)}$ \\
\textbf{} & \textbf{mrpc} & $89.7$ & $86.0_{(425.2)}$ & $87.0$ & $75.5_{(39.9)}$ \\
\textbf{} & \textbf{qnli} & $93.3$ & $93.3_{(425.2)}$ & $91.3$ & $91.3_{(39.9)}$ \\
\textbf{} & \textbf{qqp} & $91.3$ & $91.4_{(425.2)}$ & $70.4$ & $70.6_{(15.1)}$ \\
\textbf{} & \textbf{rte} & $76.2$ & $77.0_{(425.2)}$ & $74.7$ & $77.0_{(39.9)}$ \\
\textbf{} & \textbf{sst2} & $95.6$ & $93.5_{(425.2)}$ & $95.5$ & $93.5_{(31.4)}$ \\
\textbf{} & \textbf{wnli} & $56.3$ & $47.0_{(425.2)}$ & $56.3$ & $48.0_{(24.5)}$ \\
\midrule
\multirow[t]{7}{*}{\textbf{ia3}} & \textbf{mnli} & $91.0$ & $90.4_{(0.2)}$ & $90.4$ & $90.3_{(0.0)}$ \\
\textbf{} & \textbf{mrpc} & $85.5$ & $84.0_{(0.2)}$ & $85.8$ & $81.5_{(0.0)}$ \\
\textbf{} & \textbf{qnli} & $92.6$ & $92.9_{(0.2)}$ & $92.5$ & $92.4_{(0.0)}$ \\
\textbf{} & \textbf{qqp} & $89.5$ & $89.8_{(0.2)}$ & $87.1$ & $87.0_{(0.0)}$ \\
\textbf{} & \textbf{rte} & $63.9$ & $62.0_{(0.2)}$ & $65.0$ & $58.5_{(0.0)}$ \\
\textbf{} & \textbf{sst2} & $94.2$ & $93.0_{(0.2)}$ & $94.4$ & $93.0_{(0.0)}$ \\
\textbf{} & \textbf{wnli} & $52.1$ & $57.0_{(0.2)}$ & $52.1$ & $57.0_{(0.0)}$ \\
\midrule
\multirow[t]{7}{*}{\textbf{lora}} & \textbf{mnli} & $91.0$ & $90.2_{(6.2)}$ & $91.3$ & $90.5_{(0.5)}$ \\
\textbf{} & \textbf{mrpc} & $90.9$ & $84.0_{(6.2)}$ & $84.1$ & $77.5_{(0.6)}$ \\
\textbf{} & \textbf{qnli} & $93.4$ & $93.5_{(6.2)}$ & $93.3$ & $93.7_{(0.5)}$ \\
\textbf{} & \textbf{qqp} & $90.5$ & $90.5_{(6.2)}$ & $90.3$ & $90.6_{(0.4)}$ \\
\textbf{} & \textbf{rte} & $52.7$ & $53.0_{(6.2)}$ & $57.0$ & $53.5_{(0.1)}$ \\
\textbf{} & \textbf{sst2} & $94.5$ & $94.0_{(6.2)}$ & $94.4$ & $93.5_{(0.4)}$ \\
\textbf{} & \textbf{wnli} & $60.6$ & $49.0_{(6.2)}$ & $56.3$ & $47.0_{(0.1)}$ \\

\bottomrule
\end{tabular}
\end{table*}

\begin{table*}
\centering 
\caption{Validation and Test set performance along with storage size in MB for t5-large Model, for (IA)$^3$, LoRA and Full model finetuning.}
\label{tab:t5-large}
\begin{tabular}{llcccc}
\toprule
 &  & \textbf{Original (Val)} & \textbf{Original (Test)} & \textbf{ \methodshort{} (Val)} & \textbf{ \methodshort{} (Test)} \\
\textbf{PEFT} & \textbf{Task} &  &  &  &  \\
\midrule
\multirow[t]{7}{*}{\textbf{full}} & \textbf{mnli} & $93.4$ & $93.6_{(1407.0)}$ & $93.1$ & $93.2_{(81.1)}$ \\
\textbf{} & \textbf{mrpc} & $91.4$ & $88.5_{(1407.0)}$ & $90.0$ & $84.5_{(131.9)}$ \\
\textbf{} & \textbf{qnli} & $94.4$ & $94.4_{(1407.0)}$ & $94.5$ & $94.7_{(131.9)}$ \\
\textbf{} & \textbf{qqp} & $91.8$ & $91.9_{(1407.0)}$ & $68.8$ & $69.6_{(131.9)}$ \\
\textbf{} & \textbf{rte} & $83.8$ & $88.0_{(1407.0)}$ & $82.0$ & $82.0_{(131.9)}$ \\
\textbf{} & \textbf{sst2} & $93.5$ & $93.0_{(1407.0)}$ & $93.6$ & $93.0_{(131.9)}$ \\
\textbf{} & \textbf{wnli} & $56.3$ & $47.0_{(1407.0)}$ & $78.9$ & $76.0_{(103.9)}$ \\
\midrule
\multirow[t]{7}{*}{\textbf{ia3}} & \textbf{mnli} & $93.0$ & $92.5_{(0.7)}$ & $93.0$ & $92.7_{(0.1)}$ \\
\textbf{} & \textbf{mrpc} & $90.2$ & $88.5_{(0.7)}$ & $90.7$ & $90.0_{(0.0)}$ \\
\textbf{} & \textbf{qnli} & $94.4$ & $94.2_{(0.7)}$ & $94.3$ & $94.1_{(0.0)}$ \\
\textbf{} & \textbf{qqp} & $90.6$ & $91.1_{(0.7)}$ & $89.3$ & $90.0_{(0.0)}$ \\
\textbf{} & \textbf{rte} & $79.8$ & $85.0_{(0.7)}$ & $82.0$ & $83.5_{(0.0)}$ \\
\textbf{} & \textbf{sst2} & $95.5$ & $95.0_{(0.7)}$ & $95.6$ & $94.0_{(0.0)}$ \\
\textbf{} & \textbf{wnli} & $52.1$ & $57.0_{(0.7)}$ & $52.1$ & $57.0_{(0.0)}$ \\
\midrule
\multirow[t]{7}{*}{\textbf{lora}} & \textbf{mnli} & $93.0$ & $93.5_{(16.5)}$ & $93.0$ & $93.5_{(1.2)}$ \\
\textbf{} & \textbf{mrpc} & $90.9$ & $87.5_{(16.5)}$ & $85.8$ & $85.0_{(1.6)}$ \\
\textbf{} & \textbf{qnli} & $94.5$ & $94.5_{(16.5)}$ & $94.1$ & $92.9_{(1.6)}$ \\
\textbf{} & \textbf{qqp} & $90.9$ & $91.4_{(16.5)}$ & $90.2$ & $90.9_{(1.6)}$ \\
\textbf{} & \textbf{rte} & $82.0$ & $82.0_{(16.5)}$ & $78.0$ & $79.0_{(1.6)}$ \\
\textbf{} & \textbf{sst2} & $95.9$ & $95.5_{(16.5)}$ & $95.8$ & $94.0_{(1.6)}$ \\
\textbf{} & \textbf{wnli} & $56.3$ & $47.0_{(16.5)}$ & $69.0$ & $57.0_{(0.6)}$ \\

\bottomrule
\end{tabular}
\end{table*}

\begin{table*}
\centering 
\caption{Validation and Test set performance along with storage size in MB for T0-3B Model, for (IA)$^3$, and LoRA.}
\label{tab:bigscience/T0_3B}
\begin{tabular}{llcccc}
\toprule
 &  & \textbf{Original (Val)} & \textbf{Original (Test)} & \textbf{ \methodshort{} (Val)} & \textbf{ \methodshort{} (Test)} \\
\textbf{PEFT} & \textbf{Task} &  &  &  &  \\
\midrule
\multirow[t]{7}{*}{\textbf{ia3}} & \textbf{mnli} & $94.1$ & $94.4_{(1.0)}$ & $93.4$ & $93.8_{(0.1)}$ \\
\textbf{} & \textbf{mrpc} & $89.7$ & $89.5_{(1.0)}$ & $90.4$ & $89.0_{(0.1)}$ \\
\textbf{} & \textbf{qnli} & $94.9$ & $95.3_{(1.0)}$ & $95.8$ & $95.5_{(0.0)}$ \\
\textbf{} & \textbf{qqp} & $89.8$ & $90.2_{(1.0)}$ & $89.6$ & $90.0_{(0.1)}$ \\
\textbf{} & \textbf{rte} & $86.6$ & $89.0_{(1.0)}$ & $87.4$ & $88.0_{(0.0)}$ \\
\textbf{} & \textbf{sst2} & $96.8$ & $93.0_{(1.0)}$ & $96.9$ & $93.0_{(0.0)}$ \\
\textbf{} & \textbf{wnli} & $62.0$ & $74.0_{(1.0)}$ & $70.4$ & $69.0_{(0.0)}$ \\
\midrule
\multirow[t]{7}{*}{\textbf{lora}} & \textbf{mnli} & $93.8$ & $93.6_{(33.8)}$ & $93.5$ & $94.2_{(2.5)}$ \\
\textbf{} & \textbf{mrpc} & $90.4$ & $90.5_{(33.8)}$ & $90.0$ & $88.5_{(1.9)}$ \\
\textbf{} & \textbf{qnli} & $95.8$ & $94.7_{(33.8)}$ & $95.8$ & $96.0_{(2.5)}$ \\
\textbf{} & \textbf{qqp} & $90.3$ & $90.7_{(33.8)}$ & $90.4$ & $90.8_{(3.2)}$ \\
\textbf{} & \textbf{rte} & $89.2$ & $89.1_{(33.8)}$ & $88.4$ & $90.0_{(2.5)}$ \\
\textbf{} & \textbf{sst2} & $96.8$ & $95.0_{(33.8)}$ & $96.9$ & $93.0_{(2.5)}$ \\
\textbf{} & \textbf{wnli} & $73.2$ & $73.0_{(33.8)}$ & $74.6$ & $74.0_{(3.2)}$ \\

\bottomrule
\end{tabular}
\end{table*}

%% file: main.bbl
\begin{thebibliography}{91}
\providecommand{\natexlab}[1]{#1}
\providecommand{\url}[1]{\texttt{#1}}
\expandafter\ifx\csname urlstyle\endcsname\relax
  \providecommand{\doi}[1]{doi: #1}\else
  \providecommand{\doi}{doi: \begingroup \urlstyle{rm}\Url}\fi

\bibitem[Aghajanyan et~al.(2020)Aghajanyan, Zettlemoyer, and Gupta]{aghajanyan2020intrinsic}
Armen Aghajanyan, Luke Zettlemoyer, and Sonal Gupta.
\newblock Intrinsic dimensionality explains the effectiveness of language model fine-tuning.
\newblock \emph{arXiv preprint arXiv:2012.13255}, 2020.

\bibitem[Alistarh et~al.(2017)Alistarh, Grubic, Li, Tomioka, and Vojnovic]{alistarh2017qsgd}
Dan Alistarh, Demjan Grubic, Jerry Li, Ryota Tomioka, and Milan Vojnovic.
\newblock Qsgd: Communication-efficient sgd via gradient quantization and encoding.
\newblock \emph{Advances in neural information processing systems}, 30, 2017.

\bibitem[Bach et~al.(2022)Bach, Sanh, Yong, Webson, Raffel, Nayak, Sharma, Kim, Bari, Fevry, et~al.]{bach2022promptsource}
Stephen~H Bach, Victor Sanh, Zheng-Xin Yong, Albert Webson, Colin Raffel, Nihal~V Nayak, Abheesht Sharma, Taewoon Kim, M~Saiful Bari, Thibault Fevry, et~al.
\newblock Promptsource: An integrated development environment and repository for natural language prompts.
\newblock \emph{arXiv preprint arXiv:2202.01279}, 2022.

\bibitem[Bai et~al.(2022)Bai, Jones, Ndousse, Askell, Chen, DasSarma, Drain, Fort, Ganguli, Henighan, et~al.]{bai2022training}
Yuntao Bai, Andy Jones, Kamal Ndousse, Amanda Askell, Anna Chen, Nova DasSarma, Dawn Drain, Stanislav Fort, Deep Ganguli, Tom Henighan, et~al.
\newblock Training a helpful and harmless assistant with reinforcement learning from human feedback.
\newblock \emph{arXiv preprint arXiv:2204.05862}, 2022.

\bibitem[Bentivogli et~al.(2009)Bentivogli, Clark, Dagan, and Giampiccolo]{bentivogli2009rte}
Luisa Bentivogli, Peter Clark, Ido Dagan, and Danilo Giampiccolo.
\newblock The fifth pascal recognizing textual entailment challenge.
\newblock \emph{TAC}, 7:\penalty0 8, 2009.

\bibitem[Bernstein et~al.(2018)Bernstein, Wang, Azizzadenesheli, and Anandkumar]{bernstein2018signsgd}
Jeremy Bernstein, Yu-Xiang Wang, Kamyar Azizzadenesheli, and Animashree Anandkumar.
\newblock signsgd: Compressed optimisation for non-convex problems.
\newblock In \emph{International Conference on Machine Learning}, pp.\  560--569. PMLR, 2018.

\bibitem[Caccia et~al.(2023)Caccia, Ponti, Su, Pereira, Le~Roux, and Sordoni]{caccia2023multi_mhr}
Lucas Caccia, Edoardo Ponti, Zhan Su, Matheus Pereira, Nicolas Le~Roux, and Alessandro Sordoni.
\newblock Multi-head adapter routing for cross-task generalization.
\newblock In \emph{Thirty-seventh Conference on Neural Information Processing Systems}, 2023.

\bibitem[Cheng et~al.(2017)Cheng, Wang, Zhou, and Zhang]{DBLP:journals/corr/abs-1710-09282}
Yu~Cheng, Duo Wang, Pan Zhou, and Tao Zhang.
\newblock A survey of model compression and acceleration for deep neural networks.
\newblock \emph{CoRR}, abs/1710.09282, 2017.

\bibitem[Choshen et~al.(2022)Choshen, Venezian, Slonim, and Katz]{choshen2022fusing}
Leshem Choshen, Elad Venezian, Noam Slonim, and Yoav Katz.
\newblock Fusing finetuned models for better pretraining, 2022.
\newblock \url{https://arxiv.org/abs/2204.03044}.

\bibitem[Chung et~al.(2022{\natexlab{a}})Chung, Hou, Longpre, Zoph, Tay, Fedus, Li, Wang, Dehghani, Brahma, et~al.]{chung2022scaling}
Hyung~Won Chung, Le~Hou, Shayne Longpre, Barret Zoph, Yi~Tay, William Fedus, Eric Li, Xuezhi Wang, Mostafa Dehghani, Siddhartha Brahma, et~al.
\newblock Scaling instruction-finetuned language models.
\newblock \emph{arXiv preprint arXiv:2210.11416}, 2022{\natexlab{a}}.

\bibitem[Chung et~al.(2022{\natexlab{b}})Chung, Hou, Longpre, Zoph, Tay, Fedus, Li, Wang, Dehghani, Brahma, et~al.]{chung2022scaling_flant5}
Hyung~Won Chung, Le~Hou, Shayne Longpre, Barret Zoph, Yi~Tay, William Fedus, Yunxuan Li, Xuezhi Wang, Mostafa Dehghani, Siddhartha Brahma, et~al.
\newblock Scaling instruction-finetuned language models.
\newblock \emph{arXiv preprint arXiv:2210.11416}, 2022{\natexlab{b}}.

\bibitem[Dagan et~al.(2005)Dagan, Glickman, and Magnini]{dagan2005pascal}
Ido Dagan, Oren Glickman, and Bernardo Magnini.
\newblock The pascal recognising textual entailment challenge.
\newblock In \emph{Machine Learning Challenges Workshop}, 2005.
\newblock \url{https://link.springer.com/chapter/10.1007/11736790_9}.

\bibitem[Deng et~al.(2024)Deng, Zhao, Vakilian, Chen, Li, and Thrampoulidis]{deng2024dare}
Wenlong Deng, Yize Zhao, Vala Vakilian, Minghui Chen, Xiaoxiao Li, and Christos Thrampoulidis.
\newblock Dare the extreme: Revisiting delta-parameter pruning for fine-tuned models.
\newblock \emph{arXiv preprint arXiv:2410.09344}, 2024.

\bibitem[Dettmers et~al.(2023)Dettmers, Pagnoni, Holtzman, and Zettlemoyer]{dettmers2023qlora}
Tim Dettmers, Artidoro Pagnoni, Ari Holtzman, and Luke Zettlemoyer.
\newblock Qlora: Efficient finetuning of quantized llms.
\newblock \emph{arXiv preprint arXiv:2305.14314}, 2023.

\bibitem[Devlin et~al.(2018)Devlin, Chang, Lee, and Toutanova]{devlin2018bert}
Jacob Devlin, Ming-Wei Chang, Kenton Lee, and Kristina Toutanova.
\newblock {BERT}: Pre-training of deep bidirectional transformers for language understanding.
\newblock \emph{arXiv preprint arXiv:1810.04805}, 2018.

\bibitem[Dolan \& Brockett(2005)Dolan and Brockett]{dolan2005automaticallymrpc}
William~B Dolan and Chris Brockett.
\newblock Automatically constructing a corpus of sentential paraphrases.
\newblock In \emph{Proceedings of the Third International Workshop on Paraphrasing (IWP2005)}, 2005.

\bibitem[Frankle \& Carbin(2019)Frankle and Carbin]{DBLP:conf/iclr/FrankleC19}
Jonathan Frankle and Michael Carbin.
\newblock The lottery ticket hypothesis: Finding sparse, trainable neural networks.
\newblock In \emph{7th International Conference on Learning Representations}. OpenReview.net, 2019.

\bibitem[Gale et~al.(2019)Gale, Elsen, and Hooker]{DBLP:journals/corr/abs-1902-09574}
Trevor Gale, Erich Elsen, and Sara Hooker.
\newblock The state of sparsity in deep neural networks.
\newblock \emph{CoRR}, abs/1902.09574, 2019.

\bibitem[Golomb(1966)]{golomb1966run}
Solomon Golomb.
\newblock Run-length encodings (corresp.).
\newblock \emph{IEEE transactions on information theory}, 12\penalty0 (3):\penalty0 399--401, 1966.

\bibitem[Han et~al.(2015)Han, Pool, Tran, and Dally]{han2015learning}
Song Han, Jeff Pool, John Tran, and William Dally.
\newblock Learning both weights and connections for efficient neural network.
\newblock \emph{Advances in neural information processing systems}, 28, 2015.

\bibitem[Hendrycks et~al.(2020)Hendrycks, Burns, Basart, Zou, Mazeika, Song, and Steinhardt]{hendrycks2020measuring_mmlu}
Dan Hendrycks, Collin Burns, Steven Basart, Andy Zou, Mantas Mazeika, Dawn Song, and Jacob Steinhardt.
\newblock Measuring massive multitask language understanding.
\newblock \emph{arXiv preprint arXiv:2009.03300}, 2020.

\bibitem[Honovich et~al.(2022)Honovich, Scialom, Levy, and Schick]{honovich2022unnatural}
Or~Honovich, Thomas Scialom, Omer Levy, and Timo Schick.
\newblock Unnatural instructions: Tuning language models with (almost) no human labor.
\newblock \emph{arXiv preprint arXiv:2212.09689}, 2022.

\bibitem[Houlsby et~al.(2019)Houlsby, Giurgiu, Jastrzebski, Morrone, De~Laroussilhe, Gesmundo, Attariyan, and Gelly]{houlsby2019parameter_adapter}
Neil Houlsby, Andrei Giurgiu, Stanislaw Jastrzebski, Bruna Morrone, Quentin De~Laroussilhe, Andrea Gesmundo, Mona Attariyan, and Sylvain Gelly.
\newblock Parameter-efficient transfer learning for nlp.
\newblock In \emph{International Conference on Machine Learning}, pp.\  2790--2799. PMLR, 2019.

\bibitem[Hu et~al.(2021)Hu, Shen, Wallis, Allen-Zhu, Li, Wang, and Chen]{hu2021lora}
Edward~J. Hu, Yelong Shen, Phillip Wallis, Zeyuan Allen-Zhu, Yuanzhi Li, Shean Wang, and Weizhu Chen.
\newblock {LoRA}: Low-rank adaptation of large language models.
\newblock \emph{ArXiv}, abs/2106.09685, 2021.

\bibitem[Huang et~al.(2023)Huang, Liu, Lin, Pang, Du, and Lin]{huang2023lorahub}
Chengsong Huang, Qian Liu, Bill~Yuchen Lin, Tianyu Pang, Chao Du, and Min Lin.
\newblock Lorahub: Efficient cross-task generalization via dynamic lora composition.
\newblock \emph{arXiv preprint arXiv:2307.13269}, 2023.

\bibitem[Huang et~al.(2025)Huang, Ko, Zhuang, Tang, and Zhang]{huang2025hira}
Qiushi Huang, Tom Ko, Zhan Zhuang, Lilian Tang, and Yu~Zhang.
\newblock Hira: Parameter-efficient hadamard high-rank adaptation for large language models.
\newblock In \emph{The Thirteenth International Conference on Learning Representations}, 2025.

\bibitem[Ilharco et~al.(2022)Ilharco, Wortsman, Gadre, Song, Hajishirzi, Kornblith, Farhadi, and Schmidt]{ilharco2022patching}
Gabriel Ilharco, Mitchell Wortsman, Samir~Yitzhak Gadre, Shuran Song, Hannaneh Hajishirzi, Simon Kornblith, Ali Farhadi, and Ludwig Schmidt.
\newblock Patching open-vocabulary models by interpolating weights.
\newblock In \emph{Advances in Neural Information Processing Systems (NeurIPS)}, 2022.
\newblock \url{https://arXiv.org/abs/2208.05592}.

\bibitem[Ilharco et~al.(2023)Ilharco, Ribeiro, Wortsman, Schmidt, Hajishirzi, and Farhadi]{ilharco2023editing}
Gabriel Ilharco, Marco~Tulio Ribeiro, Mitchell Wortsman, Ludwig Schmidt, Hannaneh Hajishirzi, and Ali Farhadi.
\newblock Editing models with task arithmetic.
\newblock In \emph{The Eleventh International Conference on Learning Representations}, 2023.
\newblock URL \url{https://openreview.net/forum?id=6t0Kwf8-jrj}.

\bibitem[Karimi~Mahabadi et~al.(2021)Karimi~Mahabadi, Henderson, and Ruder]{karimi2021compacter}
Rabeeh Karimi~Mahabadi, James Henderson, and Sebastian Ruder.
\newblock Compacter: Efficient low-rank hypercomplex adapter layers.
\newblock \emph{Advances in Neural Information Processing Systems}, 34:\penalty0 1022--1035, 2021.

\bibitem[K{\"o}ksal et~al.(2023)K{\"o}ksal, Schick, Korhonen, and Sch{\"u}tze]{koksal2023longform}
Abdullatif K{\"o}ksal, Timo Schick, Anna Korhonen, and Hinrich Sch{\"u}tze.
\newblock Longform: Optimizing instruction tuning for long text generation with corpus extraction.
\newblock \emph{arXiv preprint arXiv:2304.08460}, 2023.

\bibitem[K{\"o}pf et~al.(2023)K{\"o}pf, Kilcher, von R{\"u}tte, Anagnostidis, Tam, Stevens, Barhoum, Duc, Stanley, Nagyfi, et~al.]{kopf2023openassistant}
Andreas K{\"o}pf, Yannic Kilcher, Dimitri von R{\"u}tte, Sotiris Anagnostidis, Zhi-Rui Tam, Keith Stevens, Abdullah Barhoum, Nguyen~Minh Duc, Oliver Stanley, Rich{\'a}rd Nagyfi, et~al.
\newblock Openassistant conversations--democratizing large language model alignment.
\newblock \emph{arXiv preprint arXiv:2304.07327}, 2023.

\bibitem[Kopiczko et~al.(2023)Kopiczko, Blankevoort, and Asano]{kopiczko2023vera}
Dawid~J Kopiczko, Tijmen Blankevoort, and Yuki~M Asano.
\newblock Vera: Vector-based random matrix adaptation.
\newblock \emph{arXiv preprint arXiv:2310.11454}, 2023.

\bibitem[LAION(2023)]{laion2023}
LAION.
\newblock Open-instruction-generalist dataset.
\newblock \url{https://github.com/LAION-AI/Open-Instruction-Generalist}, 2023.

\bibitem[Lee et~al.(2021)Lee, Park, Mo, Ahn, and Shin]{DBLP:conf/iclr/LeePMAS21}
Jaeho Lee, Sejun Park, Sangwoo Mo, Sungsoo Ahn, and Jinwoo Shin.
\newblock Layer-adaptive sparsity for the magnitude-based pruning.
\newblock In \emph{9th International Conference on Learning Representations}. OpenReview.net, 2021.

\bibitem[Lester et~al.(2021)Lester, Al-Rfou, and Constant]{lester2021prompttuning}
Brian Lester, Rami Al-Rfou, and Noah Constant.
\newblock The power of scale for parameter-efficient prompt tuning.
\newblock \emph{arXiv preprint arXiv:2104.08691}, 2021.

\bibitem[Levesque et~al.(2012{\natexlab{a}})Levesque, Davis, and Morgenstern]{levesque2012winogradwnli}
Hector Levesque, Ernest Davis, and Leora Morgenstern.
\newblock The winograd schema challenge.
\newblock In \emph{Thirteenth international conference on the principles of knowledge representation and reasoning}, 2012{\natexlab{a}}.

\bibitem[Levesque et~al.(2012{\natexlab{b}})Levesque, Davis, and Morgenstern]{wsc}
Hector Levesque, Ernest Davis, and Leora Morgenstern.
\newblock The winograd schema challenge.
\newblock \emph{Thirteenth International Conference on the Principles of Knowledge Representation and Reasoning}, 2012{\natexlab{b}}.

\bibitem[Li et~al.(2018)Li, Qian, Jiang, Lu, and Tang]{DBLP:conf/ijcai/LiQJLT18}
Guiying Li, Chao Qian, Chunhui Jiang, Xiaofen Lu, and Ke~Tang.
\newblock Optimization based layer-wise magnitude-based pruning for {DNN} compression.
\newblock In \emph{Proceedings of the Twenty-Seventh International Joint Conference on Artificial Intelligence}, pp.\  2383--2389. ijcai.org, 2018.

\bibitem[Li \& Liang(2021)Li and Liang]{li2021prefixtuning}
Xiang~Lisa Li and Percy Liang.
\newblock Prefix-tuning: Optimizing continuous prompts for generation.
\newblock \emph{arXiv preprint arXiv:2101.00190}, 2021.

\bibitem[Liang et~al.(2021)Liang, Glossner, Wang, Shi, and Zhang]{DBLP:journals/ijon/LiangGWSZ21}
Tailin Liang, John Glossner, Lei Wang, Shaobo Shi, and Xiaotong Zhang.
\newblock Pruning and quantization for deep neural network acceleration: {A} survey.
\newblock \emph{Neurocomputing}, 461:\penalty0 370--403, 2021.

\bibitem[Liu et~al.(2022)Liu, Tam, Muqeeth, Mohta, Huang, Bansal, and Raffel]{liu2022tfew}
Haokun Liu, Derek Tam, Mohammed Muqeeth, Jay Mohta, Tenghao Huang, Mohit Bansal, and Colin~A Raffel.
\newblock Few-shot parameter-efficient fine-tuning is better and cheaper than in-context learning.
\newblock \emph{Advances in Neural Information Processing Systems}, 35:\penalty0 1950--1965, 2022.

\bibitem[Liu et~al.(2024{\natexlab{a}})Liu, Xiao, Li, Lee, Han, Dao, and Cai]{liu2024bitdelta}
James Liu, Guangxuan Xiao, Kai Li, Jason~D Lee, Song Han, Tri Dao, and Tianle Cai.
\newblock Bitdelta: Your fine-tune may only be worth one bit.
\newblock \emph{Advances in Neural Information Processing Systems}, 37:\penalty0 13579--13600, 2024{\natexlab{a}}.

\bibitem[Liu et~al.(2020)Liu, Moreau, Preuss, Rapin, Roziere, Teytaud, and Teytaud]{liu2020versatile_opt}
Jialin Liu, Antoine Moreau, Mike Preuss, Jeremy Rapin, Baptiste Roziere, Fabien Teytaud, and Olivier Teytaud.
\newblock Versatile black-box optimization.
\newblock In \emph{Proceedings of the 2020 Genetic and Evolutionary Computation Conference}, pp.\  620--628, 2020.

\bibitem[Liu et~al.(2024{\natexlab{b}})Liu, Wang, Yin, Molchanov, Wang, Cheng, and Chen]{liu2024dora}
Shih-Yang Liu, Chien-Yi Wang, Hongxu Yin, Pavlo Molchanov, Yu-Chiang~Frank Wang, Kwang-Ting Cheng, and Min-Hung Chen.
\newblock Dora: Weight-decomposed low-rank adaptation.
\newblock In \emph{Forty-first International Conference on Machine Learning}, 2024{\natexlab{b}}.

\bibitem[Liu et~al.(2019{\natexlab{a}})Liu, Ott, Goyal, Du, Joshi, Chen, Levy, Lewis, Zettlemoyer, and Stoyanov]{liu2019roberta}
Yinhan Liu, Myle Ott, Naman Goyal, Jingfei Du, Mandar Joshi, Danqi Chen, Omer Levy, Mike Lewis, Luke Zettlemoyer, and Veselin Stoyanov.
\newblock Roberta: A robustly optimized bert pretraining approach, 2019{\natexlab{a}}.
\newblock \url{https://arxiv.org/abs/1907.11692}.

\bibitem[Liu et~al.(2019{\natexlab{b}})Liu, Sun, Zhou, Huang, and Darrell]{DBLP:conf/iclr/LiuSZHD19}
Zhuang Liu, Mingjie Sun, Tinghui Zhou, Gao Huang, and Trevor Darrell.
\newblock Rethinking the value of network pruning.
\newblock In \emph{7th International Conference on Learning Representations}. OpenReview.net, 2019{\natexlab{b}}.

\bibitem[Luo et~al.(2023{\natexlab{a}})Luo, Sun, Xu, Zhao, Lou, Tao, Geng, Lin, Chen, and Zhang]{luo2023wizardmath}
Haipeng Luo, Qingfeng Sun, Can Xu, Pu~Zhao, Jianguang Lou, Chongyang Tao, Xiubo Geng, Qingwei Lin, Shifeng Chen, and Dongmei Zhang.
\newblock Wizardmath: Empowering mathematical reasoning for large language models via reinforced evol-instruct.
\newblock \emph{arXiv preprint arXiv:2308.09583}, 2023{\natexlab{a}}.

\bibitem[Luo et~al.(2023{\natexlab{b}})Luo, Xu, Zhao, Sun, Geng, Hu, Tao, Ma, Lin, and Jiang]{luo2023wizardcoder}
Ziyang Luo, Can Xu, Pu~Zhao, Qingfeng Sun, Xiubo Geng, Wenxiang Hu, Chongyang Tao, Jing Ma, Qingwei Lin, and Daxin Jiang.
\newblock Wizardcoder: Empowering code large language models with evol-instruct.
\newblock \emph{arXiv preprint arXiv:2306.08568}, 2023{\natexlab{b}}.

\bibitem[Marneffe et~al.(2019)Marneffe, Simons, and Tonhauser]{cb}
Marie-Catherine~de Marneffe, Mandy Simons, and Judith Tonhauser.
\newblock {The CommitmentBank}: Investigating projection in naturally occurring discourse.
\newblock \emph{Proceedings of Sinn und Bedeutung 23}, 2019.

\bibitem[Matena \& Raffel(2021)Matena and Raffel]{matena2021merging}
Michael Matena and Colin Raffel.
\newblock Merging models with fisher-weighted averaging.
\newblock In \emph{Advances in Neural Information Processing Systems (NeurIPS)}, 2021.
\newblock \url{https://arxiv.org/abs/2111.09832}.

\bibitem[Muqeeth et~al.(2024)Muqeeth, Liu, Liu, and Raffel]{muqeeth2024phatgoose}
Mohammed Muqeeth, Haokun Liu, Yufan Liu, and Colin Raffel.
\newblock Learning to route among specialized experts for zero-shot generalization, 2024.

\bibitem[Nie et~al.(2019)Nie, Williams, Dinan, Bansal, Weston, and Kiela]{nie2019adversarial}
Yixin Nie, Adina Williams, Emily Dinan, Mohit Bansal, Jason Weston, and Douwe Kiela.
\newblock {Adversarial NLI}: A new benchmark for natural language understanding.
\newblock \emph{arXiv preprint arXiv:1910.14599}, 2019.

\bibitem[Ortiz{-}Jim{\'{e}}nez et~al.(2023)Ortiz{-}Jim{\'{e}}nez, Favero, and Frossard]{ortizjimenez2023tangent}
Guillermo Ortiz{-}Jim{\'{e}}nez, Alessandro Favero, and Pascal Frossard.
\newblock Task arithmetic in the tangent space: Improved editing of pre-trained models.
\newblock \emph{NeurIPS}, 2023.
\newblock \url{https://arxiv.org/abs/2305:12827}.

\bibitem[Pfeiffer et~al.(2023)Pfeiffer, Ruder, Vuli{\'c}, and Ponti]{pfeiffer2023modular}
Jonas Pfeiffer, Sebastian Ruder, Ivan Vuli{\'c}, and Edoardo~Maria Ponti.
\newblock Modular deep learning.
\newblock \emph{arXiv preprint arXiv:2302.11529}, 2023.

\bibitem[Pilehvar \& Camacho-Collados(2019)Pilehvar and Camacho-Collados]{pilehvar2018wic}
Mohammad~Taher Pilehvar and Jose Camacho-Collados.
\newblock {WiC}: The word-in-context dataset for evaluating context-sensitive meaning representations.
\newblock In \emph{Proceedings of NAACL-HLT}, 2019.

\bibitem[Ponti et~al.(2023)Ponti, Sordoni, Bengio, and Reddy]{ponti2023combining}
Edoardo~Maria Ponti, Alessandro Sordoni, Yoshua Bengio, and Siva Reddy.
\newblock Combining parameter-efficient modules for task-level generalisation.
\newblock In \emph{Proceedings of the 17th Conference of the European Chapter of the Association for Computational Linguistics}, pp.\  687--702, 2023.

\bibitem[Raffel et~al.(2020{\natexlab{a}})Raffel, Shazeer, Roberts, Lee, Narang, Matena, Zhou, Li, and Liu]{colin2020exploring}
Colin Raffel, Noam Shazeer, Adam Roberts, Katherine Lee, Sharan Narang, Michael Matena, Yanqi Zhou, Wei Li, and Peter~J. Liu.
\newblock Exploring the limits of transfer learning with a unified text-to-text transformer.
\newblock \emph{Journal of Machine Learning Research (JMLR)}, 2020{\natexlab{a}}.
\newblock \url{http://jmlr.org/papers/v21/20-074.html}.

\bibitem[Raffel et~al.(2020{\natexlab{b}})Raffel, Shazeer, Roberts, Lee, Narang, Matena, Zhou, Li, and Liu]{raffel2019exploring}
Colin Raffel, Noam~M. Shazeer, Adam Roberts, Katherine Lee, Sharan Narang, Michael Matena, Yanqi Zhou, Wei Li, and Peter~J. Liu.
\newblock Exploring the limits of transfer learning with a unified text-to-text transformer.
\newblock \emph{ArXiv}, abs/1910.10683, 2020{\natexlab{b}}.

\bibitem[Rajpurkar et~al.(2016)Rajpurkar, Zhang, Lopyrev, and Liang]{rajpurkar2016squadqnli}
Pranav Rajpurkar, Jian Zhang, Konstantin Lopyrev, and Percy Liang.
\newblock Squad: 100,000+ questions for machine comprehension of text.
\newblock \emph{arXiv preprint arXiv:1606.05250}, 2016.

\bibitem[Ram{\'e} et~al.(2022)Ram{\'e}, Ahuja, Zhang, Cord, Bottou, and Lopez-Paz]{rame2022modelrat}
Alexandre Ram{\'e}, Kartik Ahuja, Jianyu Zhang, Matthieu Cord, L{\'e}on Bottou, and David Lopez-Paz.
\newblock Model ratatouille: Recycling diverse models for out-of-distribution generalization.
\newblock \emph{arXiv preprint arXiv:2212.10445}, 2022.

\bibitem[Roemmele et~al.(2011)Roemmele, Bejan, and Gordon]{copa}
Melissa Roemmele, Cosmin~Adrian Bejan, and Andrew~S. Gordon.
\newblock Choice of plausible alternatives: An evaluation of commonsense causal reasoning.
\newblock \emph{2011 AAAI Spring Symposium Series}, 2011.

\bibitem[Sakaguchi et~al.(2020)Sakaguchi, Le~Bras, Bhagavatula, and Choi]{sakaguchi2020winogrande}
Keisuke Sakaguchi, Ronan Le~Bras, Chandra Bhagavatula, and Yejin Choi.
\newblock Winogrande: An adversarial winograd schema challenge at scale.
\newblock In \emph{Proceedings of the AAAI Conference on Artificial Intelligence}, 2020.

\bibitem[Sanh et~al.(2021{\natexlab{a}})Sanh, Webson, Raffel, Bach, Sutawika, Alyafeai, Chaffin, Stiegler, Scao, Raja, et~al.]{sanh2021multitask}
Victor Sanh, Albert Webson, Colin Raffel, Stephen~H. Bach, Lintang Sutawika, Zaid Alyafeai, Antoine Chaffin, Arnaud Stiegler, Teven~Le Scao, Arun Raja, et~al.
\newblock Multitask prompted training enables zero-shot task generalization.
\newblock \emph{arXiv preprint arXiv:2110.08207}, 2021{\natexlab{a}}.

\bibitem[Sanh et~al.(2021{\natexlab{b}})Sanh, Webson, Raffel, Bach, Sutawika, Alyafeai, Chaffin, Stiegler, Scao, Raja, et~al.]{sanh2022multitask}
Victor Sanh, Albert Webson, Colin Raffel, Stephen~H Bach, Lintang Sutawika, Zaid Alyafeai, Antoine Chaffin, Arnaud Stiegler, Teven~Le Scao, Arun Raja, et~al.
\newblock Multitask prompted training enables zero-shot task generalization.
\newblock In \emph{International Conference on Learning Representations (ICLR)}, 2021{\natexlab{b}}.
\newblock \url{https://arxiv.org/abs/2110.08207}.

\bibitem[Sattler et~al.(2019{\natexlab{a}})Sattler, Wiedemann, M{\"u}ller, and Samek]{sattler2019robust_stc}
Felix Sattler, Simon Wiedemann, Klaus-Robert M{\"u}ller, and Wojciech Samek.
\newblock Robust and communication-efficient federated learning from non-iid data.
\newblock \emph{IEEE transactions on neural networks and learning systems}, 31\penalty0 (9):\penalty0 3400--3413, 2019{\natexlab{a}}.

\bibitem[Sattler et~al.(2019{\natexlab{b}})Sattler, Wiedemann, M{\"u}ller, and Samek]{sattler2019sparse_sbc}
Felix Sattler, Simon Wiedemann, Klaus-Robert M{\"u}ller, and Wojciech Samek.
\newblock Sparse binary compression: Towards distributed deep learning with minimal communication.
\newblock In \emph{2019 International Joint Conference on Neural Networks (IJCNN)}, pp.\  1--8. IEEE, 2019{\natexlab{b}}.

\bibitem[Sharma et~al.(2018)Sharma, Allen, Bakhshandeh, and Mostafazadeh]{sharma2018tackling}
Rishi Sharma, James Allen, Omid Bakhshandeh, and Nasrin Mostafazadeh.
\newblock Tackling the story ending biases in the story cloze test.
\newblock In \emph{Proceedings of the 56th Annual Meeting of the Association for Computational Linguistics (Volume 2: Short Papers)}, pp.\  752--757, 2018.

\bibitem[Sheng et~al.(2023)Sheng, Cao, Li, Hooper, Lee, Yang, Chou, Zhu, Zheng, Keutzer, Gonzalez, and Stoica]{sheng2023slora}
Ying Sheng, Shiyi Cao, Dacheng Li, Coleman Hooper, Nicholas Lee, Shuo Yang, Christopher Chou, Banghua Zhu, Lianmin Zheng, Kurt Keutzer, Joseph~E. Gonzalez, and Ion Stoica.
\newblock S-lora: Serving thousands of concurrent lora adapters.
\newblock \emph{arXiv preprint arXiv:2311.03285}, 2023.

\bibitem[Socher et~al.(2013)Socher, Perelygin, Wu, Chuang, Manning, Ng, and Potts]{socher2013sst2}
Richard Socher, Alex Perelygin, Jean Wu, Jason Chuang, Christopher~D Manning, Andrew~Y Ng, and Christopher Potts.
\newblock Recursive deep models for semantic compositionality over a sentiment treebank.
\newblock In \emph{Proceedings of the 2013 conference on empirical methods in natural language processing}, pp.\  1631--1642, 2013.

\bibitem[Strom(2015)]{strom2015scalable}
Nikko Strom.
\newblock Scalable distributed dnn training using commodity gpu cloud computing.
\newblock In \emph{Interspeech}, 2015.
\newblock URL \url{https://api.semanticscholar.org/CorpusID:9338808}.

\bibitem[Suzgun et~al.(2022)Suzgun, Scales, Sch{\"a}rli, Gehrmann, Tay, Chung, Chowdhery, Le, Chi, Zhou, , and Wei]{suzgun2022challenging_bbh}
Mirac Suzgun, Nathan Scales, Nathanael Sch{\"a}rli, Sebastian Gehrmann, Yi~Tay, Hyung~Won Chung, Aakanksha Chowdhery, Quoc~V Le, Ed~H Chi, Denny Zhou, , and Jason Wei.
\newblock Challenging big-bench tasks and whether chain-of-thought can solve them.
\newblock \emph{arXiv preprint arXiv:2210.09261}, 2022.

\bibitem[Taori et~al.(2023)Taori, Gulrajani, Zhang, Dubois, Li, Guestrin, Liang, and Hashimoto]{alpaca}
Rohan Taori, Ishaan Gulrajani, Tianyi Zhang, Yann Dubois, Xuechen Li, Carlos Guestrin, Percy Liang, and Tatsunori~B. Hashimoto.
\newblock Stanford alpaca: An instruction-following llama model.
\newblock \url{https://github.com/tatsu-lab/stanford_alpaca}, 2023.

\bibitem[Touvron et~al.(2023{\natexlab{a}})Touvron, Lavril, Izacard, Martinet, Lachaux, Lacroix, Rozi{\`e}re, Goyal, Hambro, Azhar, et~al.]{touvron2023llama}
Hugo Touvron, Thibaut Lavril, Gautier Izacard, Xavier Martinet, Marie-Anne Lachaux, Timoth{\'e}e Lacroix, Baptiste Rozi{\`e}re, Naman Goyal, Eric Hambro, Faisal Azhar, et~al.
\newblock Llama: Open and efficient foundation language models.
\newblock \emph{arXiv preprint arXiv:2302.13971}, 2023{\natexlab{a}}.

\bibitem[Touvron et~al.(2023{\natexlab{b}})Touvron, Martin, Stone, Albert, Almahairi, Babaei, Bashlykov, Batra, Bhargava, Bhosale, et~al.]{touvron2023llama2}
Hugo Touvron, Louis Martin, Kevin Stone, Peter Albert, Amjad Almahairi, Yasmine Babaei, Nikolay Bashlykov, Soumya Batra, Prajjwal Bhargava, Shruti Bhosale, et~al.
\newblock Llama 2: Open foundation and fine-tuned chat models.
\newblock \emph{arXiv preprint arXiv:2307.09288}, 2023{\natexlab{b}}.

\bibitem[Wang et~al.(2018{\natexlab{a}})Wang, Singh, Michael, Hill, Levy, and Bowman]{wang2018glue}
Alex Wang, Amanpreet Singh, Julian Michael, Felix Hill, Omer Levy, and Samuel~R Bowman.
\newblock Glue: A multi-task benchmark and analysis platform for natural language understanding.
\newblock \emph{EMNLP 2018}, pp.\  353, 2018{\natexlab{a}}.

\bibitem[Wang et~al.(2018{\natexlab{b}})Wang, Sievert, Liu, Charles, Papailiopoulos, and Wright]{wang2018atomo}
Hongyi Wang, Scott Sievert, Shengchao Liu, Zachary Charles, Dimitris Papailiopoulos, and Stephen Wright.
\newblock Atomo: Communication-efficient learning via atomic sparsification.
\newblock \emph{Advances in neural information processing systems}, 31, 2018{\natexlab{b}}.

\bibitem[Wang et~al.(2022)Wang, Kordi, Mishra, Liu, Smith, Khashabi, and Hajishirzi]{wang2022self}
Yizhong Wang, Yeganeh Kordi, Swaroop Mishra, Alisa Liu, Noah~A Smith, Daniel Khashabi, and Hannaneh Hajishirzi.
\newblock Self-instruct: Aligning language model with self generated instructions.
\newblock \emph{arXiv preprint arXiv:2212.10560}, 2022.

\bibitem[Wen et~al.(2017)Wen, Xu, Yan, Wu, Wang, Chen, and Li]{wen2017terngrad}
Wei Wen, Cong Xu, Feng Yan, Chunpeng Wu, Yandan Wang, Yiran Chen, and Hai Li.
\newblock Terngrad: Ternary gradients to reduce communication in distributed deep learning.
\newblock \emph{Advances in neural information processing systems}, 30, 2017.

\bibitem[Williams et~al.(2018)Williams, Nangia, and Bowman]{williams2018mnli}
Adina Williams, Nikita Nangia, and Samuel~R Bowman.
\newblock A broad-coverage challenge corpus for sentence understanding through inference.
\newblock In \emph{Proceedings of NAACL-HLT}, pp.\  1112--1122, 2018.

\bibitem[Wolf et~al.(2019)Wolf, Debut, Sanh, Chaumond, Delangue, Moi, Cistac, Rault, Louf, Funtowicz, et~al.]{wolf2019huggingface}
Thomas Wolf, Lysandre Debut, Victor Sanh, Julien Chaumond, Clement Delangue, Anthony Moi, Pierric Cistac, Tim Rault, R{\'e}mi Louf, Morgan Funtowicz, et~al.
\newblock Huggingface's transformers: State-of-the-art natural language processing, 2019.
\newblock \url{https://arxiv.org/abs/1910.03771}.

\bibitem[Wortsman et~al.(2022{\natexlab{a}})Wortsman, Ilharco, Gadre, Roelofs, Gontijo-Lopes, Morcos, Namkoong, Farhadi, Carmon, Kornblith, et~al.]{wortsman2022model}
Mitchell Wortsman, Gabriel Ilharco, Samir~Yitzhak Gadre, Rebecca Roelofs, Raphael Gontijo-Lopes, Ari~S Morcos, Hongseok Namkoong, Ali Farhadi, Yair Carmon, Simon Kornblith, et~al.
\newblock Model soups: averaging weights of multiple fine-tuned models improves accuracy without increasing inference time.
\newblock In \emph{International Conference on Machine Learning (ICML)}, 2022{\natexlab{a}}.
\newblock \url{https://arxiv.org/abs/2203.05482}.

\bibitem[Wortsman et~al.(2022{\natexlab{b}})Wortsman, Ilharco, Li, Kim, Hajishirzi, Farhadi, Namkoong, and Schmidt]{wortsman2021robust}
Mitchell Wortsman, Gabriel Ilharco, Mike Li, Jong~Wook Kim, Hannaneh Hajishirzi, Ali Farhadi, Hongseok Namkoong, and Ludwig Schmidt.
\newblock Robust fine-tuning of zero-shot models.
\newblock In \emph{Conference on Computer Vision and Pattern Recognition (CVPR)}, 2022{\natexlab{b}}.
\newblock \url{https://arxiv.org/abs/2109.01903}.

\bibitem[Xia et~al.(2022)Xia, Zhong, and Chen]{DBLP:conf/acl/XiaZC22}
Mengzhou Xia, Zexuan Zhong, and Danqi Chen.
\newblock Structured pruning learns compact and accurate models.
\newblock In Smaranda Muresan, Preslav Nakov, and Aline Villavicencio (eds.), \emph{Proceedings of the 60th Annual Meeting of the Association for Computational Linguistics (Volume 1: Long Papers)}, pp.\  1513--1528. Association for Computational Linguistics, 2022.

\bibitem[Yadav et~al.(2023)Yadav, Tam, Choshen, Raffel, and Bansal]{yadav2023ties-merging}
Prateek Yadav, Derek Tam, Leshem Choshen, Colin Raffel, and Mohit Bansal.
\newblock Ties-merging: Resolving interference when merging models.
\newblock In \emph{Thirty-seventh Conference on Neural Information Processing Systems}, 2023.

\bibitem[Yadav et~al.(2024)Yadav, Raffel, Muqeeth, Caccia, Liu, Chen, Bansal, Choshen, and Sordoni]{yadav2024survey}
Prateek Yadav, Colin Raffel, Mohammed Muqeeth, Lucas Caccia, Haokun Liu, Tianlong Chen, Mohit Bansal, Leshem Choshen, and Alessandro Sordoni.
\newblock A survey on model moerging: Recycling and routing among specialized experts for collaborative learning.
\newblock \emph{CoRR}, 2024.

\bibitem[Yang et~al.(2023)Yang, Li, Zhang, and Zong]{yang2023bigtrans}
Wen Yang, Chong Li, Jiajun Zhang, and Chengqing Zong.
\newblock Bigtrans: Augmenting large language models with multilingual translation capability over 100 languages.
\newblock \emph{arXiv preprint arXiv:2305.18098}, 2023.

\bibitem[Yu et~al.(2023)Yu, Yu, Yu, Huang, and Li]{yu2023mario_merging}
Le~Yu, Bowen Yu, Haiyang Yu, Fei Huang, and Yongbin Li.
\newblock Language models are super mario: Absorbing abilities from homologous models as a free lunch.
\newblock \emph{arXiv preprint arXiv:2311.03099}, 2023.

\bibitem[Zaken et~al.(2021)Zaken, Ravfogel, and Goldberg]{zaken2021bitfit}
Elad~Ben Zaken, Shauli Ravfogel, and Yoav Goldberg.
\newblock Bitfit: Simple parameter-efficient fine-tuning for transformer-based masked language-models.
\newblock \emph{arXiv preprint arXiv:2106.10199}, 2021.

\bibitem[Zellers et~al.(2019)Zellers, Holtzman, Bisk, Farhadi, and Choi]{zellers2019hellaswag}
Rowan Zellers, Ari Holtzman, Yonatan Bisk, Ali Farhadi, and Yejin Choi.
\newblock {HellaSwag}: Can a machine really finish your sentence?
\newblock \emph{arXiv preprint arXiv:1905.07830}, 2019.

\bibitem[Zhang et~al.(2023)Zhang, Han, Zhou, Hu, Yan, Lu, Li, Gao, and Qiao]{zhang2023llama_adapter}
Renrui Zhang, Jiaming Han, Aojun Zhou, Xiangfei Hu, Shilin Yan, Pan Lu, Hongsheng Li, Peng Gao, and Yu~Qiao.
\newblock Llama-adapter: Efficient fine-tuning of language models with zero-init attention.
\newblock \emph{arXiv preprint arXiv:2303.16199}, 2023.

\bibitem[Zhu \& Gupta(2018)Zhu and Gupta]{DBLP:conf/iclr/ZhuG18}
Michael Zhu and Suyog Gupta.
\newblock To prune, or not to prune: Exploring the efficacy of pruning for model compression.
\newblock In \emph{6th International Conference on Learning Representations}. OpenReview.net, 2018.

\end{thebibliography}
